\documentclass[runningheads]{llncs}

\usepackage{eccv}

\usepackage{eccvabbrv}

\usepackage{graphicx}
\usepackage{booktabs}
\usepackage{tabularx}
\usepackage{wrapfig}
\usepackage{tablefootnote}

\usepackage{tikz}
\newcommand*\circled[1]{\tikz[baseline=(char.base)]{
            \node[shape=circle,draw,inner sep=2pt] (char) {#1};}}
\usepackage[accsupp]{axessibility}  

\usepackage{hyperref}

\usepackage{orcidlink}
\usepackage{multirow}
\usepackage{xparse}
\usepackage{xcolor,colortbl}

\usepackage{amssymb}
\usepackage{pifont}

\newcommand*\colourcheck[1]{%
  \expandafter\newcommand\csname #1check\endcsname{\textcolor{#1}{\ding{51}}}%
}
\colourcheck{green}
\newcommand*\colourx[1]{%
  \expandafter\newcommand\csname #1x\endcsname{\textcolor{#1}{\ding{55}}}%
}
\colourx{red}

\newcommand{\cmark}{\greencheck}%
\newcommand{\xmark}{\redx}%

\newcommand{\name}{\textit{m}\&\textit{m}'s\xspace}
\newcommand{\modelnum}{10 }
\newcommand{\tool}[1]{\texttt{#1}}
\newcommand{\node}[2]{$\langle node-#1\rangle$.#2}
\definecolor{amber}{rgb}{1.0, 0.75, 0.0}

\begin{document}

\title{\parbox{0.05\textwidth}{\includegraphics[width=\linewidth]{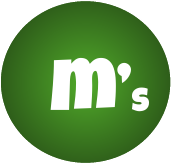}} \name: A Benchmark to Evaluate Tool-Use for  \textit{m}ulti-step \textit{m}ulti-modal Tasks}  

\titlerunning{m\&ms}


\author{Zixian Ma\inst{1}, Weikai Huang \inst{1},
Jieyu Zhang\inst{1}, Tanmay Gupta\inst{2}, Ranjay Krishna\inst{1,2}}

\authorrunning{~Ma et al.}

\institute{University of Washington \and
Allen Institute of Artificial Intelligence
\\}

\maketitle
\begin{abstract}
Real-world multi-modal problems are rarely solved by a single machine learning model, and often require multi-step computational plans that involve stitching several models. Tool-augmented LLMs hold tremendous promise for automating the generation of such computational plans. However, the lack of standardized benchmarks for evaluating LLMs as planners for multi-step multi-modal tasks has prevented a systematic study of planner design decisions.
Should LLMs generate a full plan in a single shot or step-by-step? Should they invoke tools directly with Python code or through structured data formats like JSON? Does feedback improve planning? 
To answer these questions and more, we introduce \name: a benchmark containing 4K+ \textit{m}ulti-step \textit{m}ulti-modal tasks involving 33 tools that include multi-modal models, (free) public APIs, and image processing modules. For each of these task queries, we provide automatically generated plans using this realistic toolset. We further provide a high-quality subset of 1,565 task plans that are human-verified and correctly executable. With \name, we evaluate \modelnum\ popular LLMs with 2 planning strategies (multi-step vs.~step-by-step planning), 2 plan formats (JSON vs.~code), and 3 types of feedback (parsing/verification/execution). Finally, we summarize takeaways from our extensive experiments and provide practical recommendations for designing planners for multi-step multi-modal tasks. Our dataset and evaluation code are available on HuggingFace\footnote{https://huggingface.co/datasets/zixianma/mms} and Github\footnote{https://github.com/RAIVNLab/mms} respectively.

\end{abstract}
\section{Introduction}
\begin{figure}
    \centering
    \includegraphics[width=\textwidth]{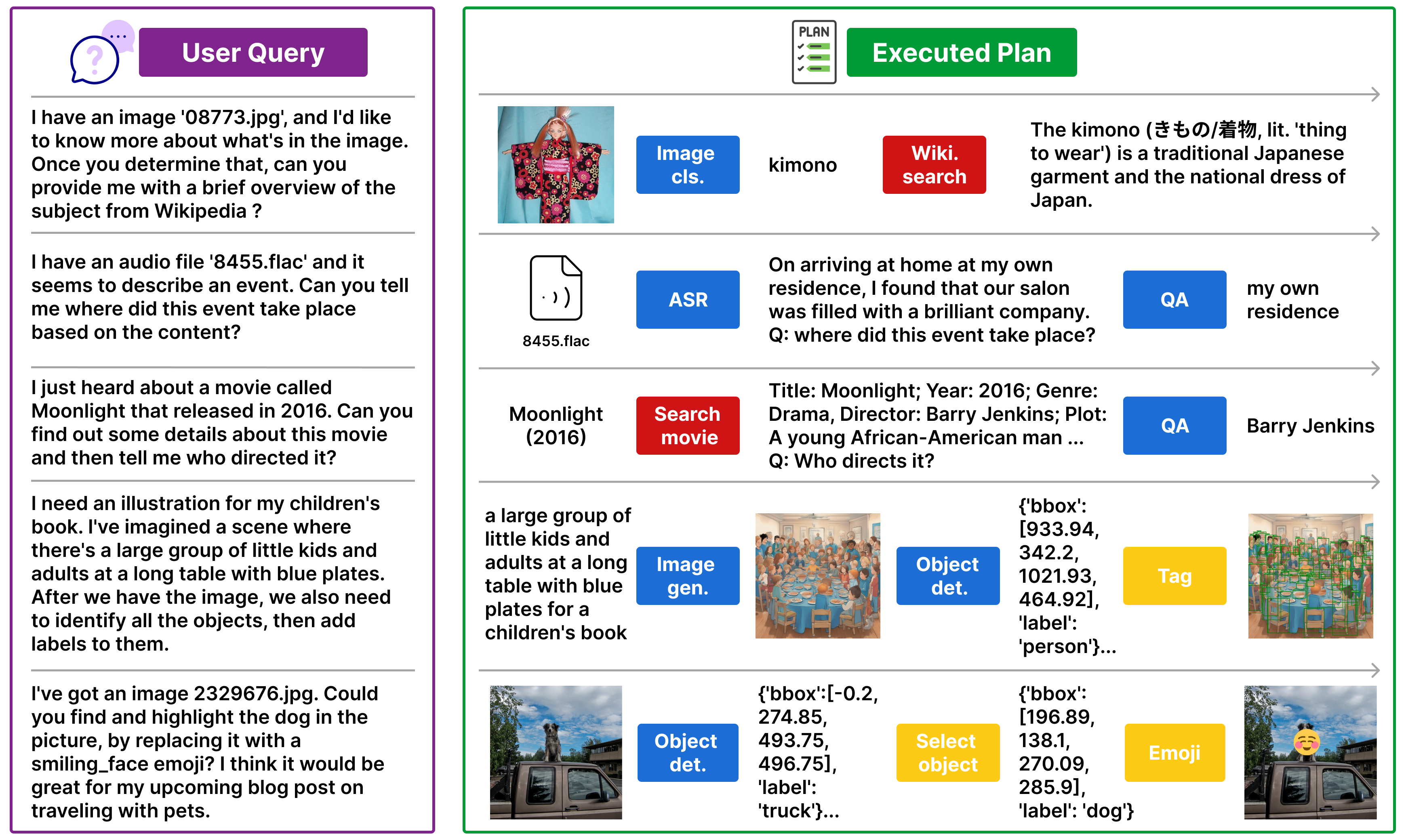}
    \caption{We present examples of query-plan pairs along with the execution results of the plans in \name. Our benchmark contains a large quantity of diverse user queries involving three modalities (i.e. text, image, and audio) as well as human-verified plans that consist of 1 - 3 tools across three categories: \textcolor{blue}{multi-modal machine learning models} (blue), \textcolor{red}{public APIs} (red) and \textcolor{amber}{image processing modules} (yellow).}
    \label{fig:dataset-examples}
\end{figure}

Planning agents—powered by large language models (LLMs)—are becoming increasingly proficient at decomposing user-specified tasks into a series of subtasks, where each subtask is executed by invoking tools. For example, if a user wants to learn about a new piece of clothing in an image, the LLM planner can create a plan with multiple steps: first, it invokes an image classification tool to identify the clothing as a ``kimono''; once identified, it can issue a Wikipedia search query to explain the cultural relevance of the kimono (Figure~\ref{fig:dataset-examples} first row).  

LLM planning agents typically consist of an LLM and a set of tools to plan over. 
Given an LLM and toolset, the design space of planning agents is extremely rich, involving many decisions such as \mbox{\textit{planning strategy}} (e.g. generation of the whole plan vs one step of the plan at a time), \mbox{\textit{forms of feedback}} (e.g. no feedback or parsing/verification/execution feedback), and \mbox{\textit{plan format}} (e.g. JSON strings that specify tools and arguments vs free-form Python code). 

Unfortunately, there is no existing planning benchmark that supports evaluation along this combinatorially rich design space with a realistic set of multimodal tools. Recent concurrent benchmarks such as ToolEmu and TaskBench~\cite{ruan2023toolemu, shen2023taskbench} provide user queries and ground truth plans but lack support for realistic plan execution. For instance, TaskBench assumes that a list of tools is available for planning without providing implementation of them. TaskBench also does not instantiate the queries with actual inputs and uses placeholder input filenames like “example.png” that do not exist. ToolEmu likewise uses LLMs to emulate tool execution instead of providing tool implementations. The lack of actual implementations of tools and real execution feedback while planning makes the study of the design space elucidated above unrealistic at best, if not impossible. 

Motivated by this dire need for a standardized benchmark for studying the design space of multi-step multi-modal planning agents, we first propose the \name benchmark. \name contains 4K+ realistic user tasks and automatically generated task plans. 1565 of these task plans are human-verified and executable with 33 curated tools consisting of multi-modal models, (free) public APIs, and image processing modules. 

Next, we use \name to systematically study the impact of 2 planning strategies (step-by-step and multi-step), 3 kinds of feedback (parsing, verification and execution), and 2 plan formats (JSON and code). Through extensive experimentation with 10 LLMs -- 5 popular open-source LLMs, 2 code LLMs and 3 proprietary LLMs of varying sizes -- we provide a series of findings:

First, existing LLMs instructed to perform multi-step planning consistently outperform step-by-step planning on \name tasks, although the performance gap is smaller with more capable larger models such as Llama-3-70B and GPT-4.
Second, verification and execution feedback improve LLMs' ability to generate overall executable plans and predict the correct argument names but don't necessarily improve their tool selection ability. We also observe a smaller improvement from verification/execution feedback on larger models such as Llama-3-70B and GPT-4, which already obtain relatively high scores with only parsing feedback.
Third, LLMs perform comparably on tool prediction with JSON-format and Python code generation, but most models produce more executable plans with JSON-format generation. Nonetheless, this gap in executability is smaller for code LLMs such as CodeLlama-34B and 70B.
Taken together, our experiments suggest that for multi-step multi-modal tasks, multi-step planning in JSON with feedback can result in the best overall tool-use performance compared to step-by-step planning, code generation, or the same setup without feedback. 

\section{Related work}
\begingroup
\begin{table}[t]
\centering
\caption{Compared to previous tool planning benchmarks, \name contains multimodal queries that are more realistic and executable. *: MetaTool only considers Open AI plugins as tools. \#: The queries of TaskBench contain textural placeholder of other modality data such as images, while queries of \name come with real images.}
\label{tab:benchmark_comparison}
 \resizebox{0.9\textwidth}{!}{%
\footnotesize
\begin{tabular}{llccccc}
\toprule
 &  & ToolBench & ToolEmu & TaskBench & MetaTool & \name  \\ 
 &  & \cite{qin2023toolllm} & \cite{ruan2023toolemu} & \cite{shen2023taskbench} & \cite{huang2023metatool} & (ours) \\ 
\cmidrule{1-7}\morecmidrules\cmidrule{1-7}

 \multirow{2}{*}{Query} & Real multi-modal inputs? & \xmark & \xmark & \xmark$^{\#}$ & \xmark &\cmark \\ 
  & {Verified by human?} & \xmark & \cmark & \cmark & \cmark & \cmark \\ \hline 

\multirow{2}{*}{Tools} & {Are all tools executable?} & \cmark & \xmark & \xmark & \cmark & \cmark \\
 & Multi-modal models & \xmark & \xmark & \cmark & \xmark$^{*}$ & \cmark \\ 
 \hline


Plan & Format & JSON & JSON & JSON & JSON & JSON/Code \\ \hline
\multirow{2}{*}{Scale} & Number of unique tools & 3,451 & 36 & 103 & 390 & 33 \\ 
 & Number of queries & 126k & 144 & 17K & 20k & 1.5k \\ 
\bottomrule
\end{tabular}%
}

\end{table}
\endgroup
We situate our work amongst the ever-growing number of tool-use research. 
\noindent\textbf{Planning evaluations.}
Although many tool-use variants have been proposed, evaluating LLMs on tool-use still lacks a standardized protocol.
For instance, VisProg and ViperGPT evaluate their plan's \textit{executions} on vision tasks using a Python-like \textit{code} format~\cite{gupta2022visual,suris2023vipergpt}.
HuggingGPT evaluates only the \textit{plan} accuracy (did the agent choose the right tools) without executing the proposed plans~\cite{shen2023hugginggpt}.
ToolFormer~\cite{schick2024toolformer} and ToolLLaMA~\cite{qin2023toolllm} both use \textit{natural language} instead of \textit{code} to interface with tools; while ToolFormer generates a \textit{multi-step} plan all at once and evaluates the program's \textit{execution}, ToolLLaMA generates the plan \textit{step-by-step}, with \textit{self-feedback} to correct mistakes. ToolLLaMA evaluates only the \textit{plans} while ToolFormer evaluates both \textit{plans} and executions.
Unfortunately, no single benchmark evaluates planning agents along this combinatorial design space, which is what we contribute.

\noindent\textbf{Tool-use benchmarks.}
Today, tool-use evaluation is spread out across a number of diverse benchmarks, including HotpotQA, WebShop, GQA, RefCOCO, and NLVR~\cite{yang-etal-2018-hotpotqa,yao2022webshop,hudson2019gqa,kazemzadeh-etal-2014-referitgame,suhr-etal-2017-corpus}. None of these contains ground truth plans, conflating planning errors with execution error. In other words, it is hard to separate whether an LLM failed to propose the correct plan or whether one of the tools used in the plan failed.
In response, recent concurrent efforts have proposed new benchmarks, such as ToolEmu, TaskBench, and GAIA~\cite{ruan2023toolemu, shen2023taskbench, mialon2023gaia}. They do contain ground truth plans but fail to support evaluating plans' execution results. 

\noindent\textbf{Planning strategies.}
There are multiple strategies for planning. For instance, Psychology literature reveals that people rarely plan tasks in their entirety due to the cognitive cost of planning long-range tasks \cite{correa2023humans}. Instead, they plan the first couple of subtasks, and execute them before planning the rest~\cite{correa2023humans,allen2020rapid}. 
In the tool-use literature, we identify two primary forms of planning strategies: \emph{step-by-step planning}~\cite{yao2023react,qin2023toolllm,gao2023assistgpt} and \emph{multi-step planning}~\cite{gupta2022visual,suris2023vipergpt,shen2023hugginggpt}.
Similar to people, step-by-step planning generates plans sequentially with one subtask at a time. 
By contrast, multi-step planning creates the entire plan before executing any subtask.
Unfortunately, these two strategies have not been systematically compared; we systematically compare both across multiple open-source and close-source LLMs.

\noindent\textbf{Feedback mechanisms.}
LLM planners make mistakes, stitching together tools that fail to execute or worse, fail to compile. 
Although human feedback is one mechanism to align plans with human expectations and preferences~\cite{chen2023interact,wang2023mint}, they require real users, making evaluation stochastic. However, there have been several automatic mechanisms that can improve plans~\cite{zhang2023ecoassistant,wang2023survey}.
For instance, syntactic mistakes can easily be detected using external \textit{verifiers} and can guide planners to iterate on their plans~\cite{shinn2023reflexion,huang2022inner,madaan2024self, miao2023selfcheck}.
Others require examining the output of individual subtask \textit{executions}~\cite{yao2023react,wang2023voyager, zhu2023ghost, rana2023sayplan,sun2024adaplanner}.
In this work, we compare plan parsing/verification feedback as well as tool execution feedback. 

\section{\name: the benchmark}
To facilitate the study of LLM planners for \textit{m}ulti-step \textit{m}ulti-modal tasks, we curate the \name benchmark. 
Before describing the dataset generation process, we first formalize the tool-planning problem in Sec~\ref{sec:problem}.  
We then describe our benchmark creation process in Sec.~\ref{sec:data_gen} and present dataset statistics in Sec.~\ref{sec:data_stats}.

\subsection{Formalizing multi-step multi-modal tool-use}
\label{sec:problem}
Given a tool set $\mathcal{T}$, and the user query $\mathcal{Q}$, a planner is required to produce a plan $\mathcal{P}$ that consists of a sequence of tool invocations $\mathcal{P}=[t_1(\{a^k_1=v^k_1\}_k), t_2(\{a^k_2=v^k_2\}_k), \cdots, t_m(\{a^k_m=v^k_m\}_k)]$, where $t_j$ represents the $j^{th}$ tool in the plan, and $a^{k}_{j}$, and $v^{k}_{j}$ represent tool $t_j$'s $k^{th}$ argument name and value respectively. Note that the output of $t_j$ may be used as argument values for subsequent tools $t_{j+1:m}$. \name contains a set of $N$ query-plan pairs, i.e., $\{(\mathcal{Q}_{i}, \mathcal{P}_{i})\}_{i\in [N]}$ with each plan composed of executable tools chosen from a curated set of API-calling functions, multi-modal models, and image processing modules. 


\subsection{Dataset generation}~\label{sec:data_gen}
\begin{figure}[t]
    \centering
    \includegraphics[width=0.9\textwidth]{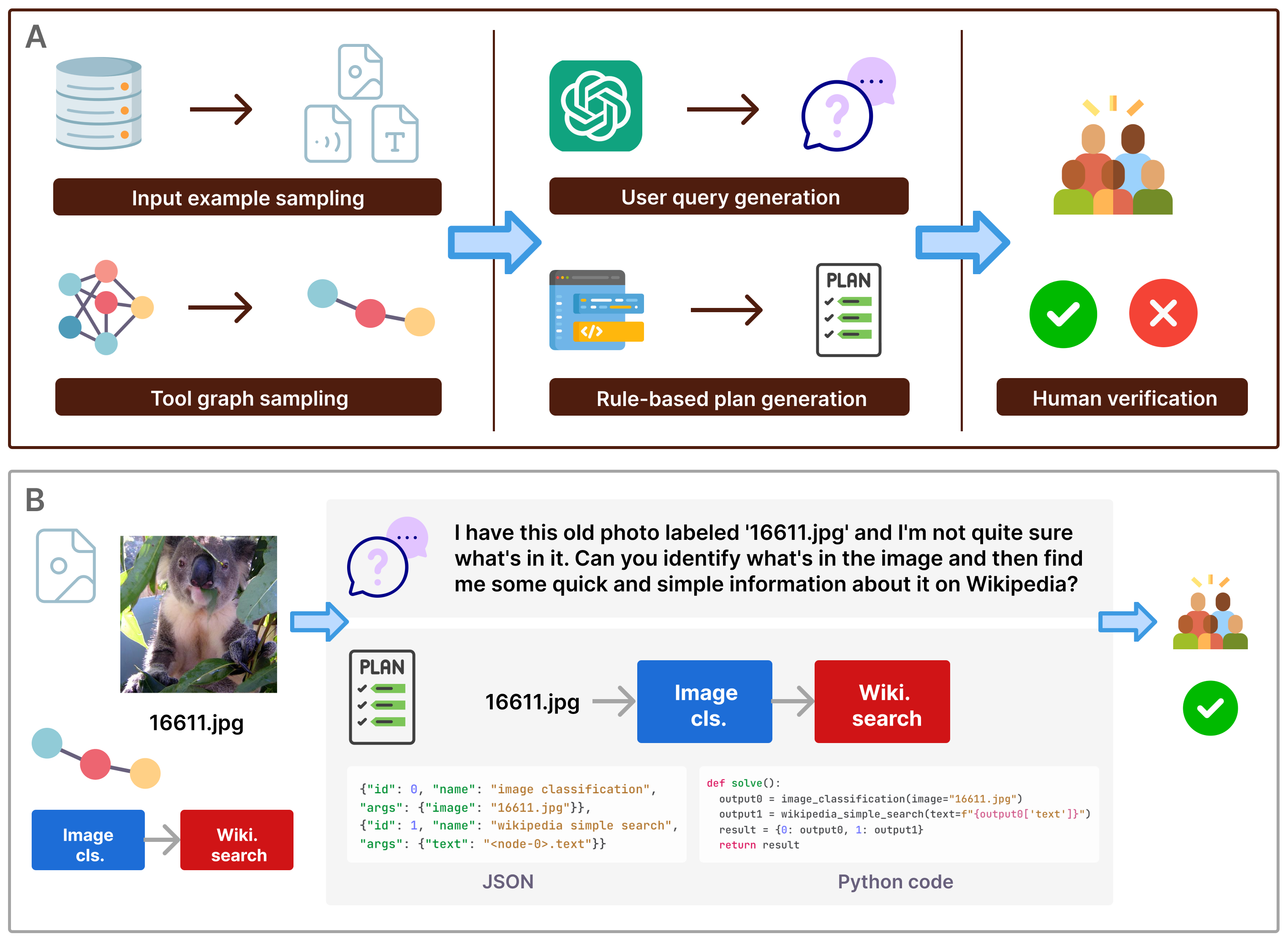}
    \caption{\textbf{Data generation pipeline.} As shown in panel A, our generation process consists of five main steps: input example sampling, tool graph sampling, user query generation with GPT-4, rule-based plan generation, and human verification. Panel B showcases an instantiation of this process with a real example.}
    \label{fig:dataset-gen}
\end{figure}

To create such a dataset, our data generation process consists of five major steps:
\circled{1} We construct a valid tool graph with all the tools and sample a subgraph from it as a sequence of tools in the target plan;
\circled{2} To instantiate the tool plan with inputs, we pair plans with real inputs (e.g., images) from existing datasets based on the first tool in the plan;
\circled{3} To generate the user query, we prompt GPT-4 with the tool graph and input pair to generate realistic user queries with few-shot demonstrations; 
\circled{4} We transform the tool graph and input pair to a fully specified JSON plan (tool names with corresponding argument names and values) with a rule-based program. Additionally, we map the JSON-format plans to Python code to support code generation evaluation;
\circled{5} Finally, three human annotators verify whether each plan can solve the corresponding user query. To obtain the final tool-balanced human-verified subset, we discard some plans from the initial human-verified set to avoid an overwhelming representation of any tool.

\noindent\circled{1} \textbf{Tool graph sampling.} 
We first create a directed graph with all 33 tools as the nodes and edges denoting valid connections between nodes. 
A connection is valid only if the output of the source tool matches the expected input type of the target tool. For example, there is an edge between \tool{image classification} and \tool{wikipedia simple search}, because the output of \tool{image classification} - a text label - is a valid input type for \tool{wikipedia simple search}.   We then sample subgraphs from the full tool graph to obtain tool sequences with valid tool dependencies. 

\noindent\circled{2} \textbf{Input example sampling.} We now need to instantiate queries with real input examples. To do so, we first collect real-world examples from the validation sets of 11 existing datasets, including ImageNet~\cite{deng2009imagenet}, SQUAD~\cite{rajpurkar-etal-2016-squad}, Visual Genome~\cite{krishna2017visual}, MagicBrush~\cite{zhang2024magicbrush}, librispeech~\cite{7178964}. Then, to pair a tool graph sampled in the previous step with an input, we randomly sample an input based on the input type needed for the first tool in the graph. For example, if the first tool in a tool sequence is \tool{image classification}, we randomly sample an image (e.g. ``16611.jpg'') from ImageNet as the input. 

\noindent\circled{3} \textbf{Query generation.} With a set of tool sequences and input examples to the first tools, we prompt GPT-4 to generate realistic user queries. Concretely, we randomly sample five different input examples for each tool sequence and ask GPT-4 to generate two queries for each tool sequence with the same input (See Appendix for the full prompt). 

\noindent\circled{4} \textbf{Plan generation.} For plan generation, we write a rule-based program to generate a plan (i.e. an ordered list of tool names with corresponding argument names and values fully specified) for each query.  Each step in the plan contains an id, tool name, and an argument dictionary with this tool's argument names as the keys and argument values as values. We populate each node's ID and name based on the sampled tool sequence and fill in the argument names for each tool using a pre-defined metadata document. We also fill in the argument values of the first tool using the input examples and those of subsequent tools using a special notation \node{id}{key}, where id refers to the id of the previous node and key refers to the output key. To further refine the plans to be even more faithful to queries, we rewrite the argument values of \tool{text generation} and \tool{image generation} (e.g. from ``a shark'' to ``a child-friendly book cover image of a shark'') by prompting GPT-4 with the queries and original plans. 

\noindent\circled{5} \textbf{Human verification} Finally, we perform extensive human verification on all 4427 generated query-plan pairs. We ask three expert annotators (who are undergraduate and Ph.D. students in CS) to rate each query-plan pair with 0 or 1, where 1 indicates that the plan can resolve the query perfectly. We obtain a subset of 1500+ examples on which all three annotators rate 1 and perform further filtering of examples where the plan contains much more frequent tools (e.g. \tool{image generation} and \tool{text generation}) to balance the overall distribution of tools (See Appendix for more details on filtering and the tool distribution). 

It is worth noting that two of the steps in our dataset generation pipeline draw similarities with the recently released concurrent TaskBench~\cite{shen2023taskbench}. Similar to them, we also sample a subgraph of tools and query generation steps. However, we want to highlight two major differences: first, we leverage real-world examples as inputs to the tool sequences (in contrast to TaskBench's ``example.jpg'', ``example.wav'' etc.), which not only leads to a more realistic instantiation of queries but also enables plan execution on actual input which is crucial for studying execution feedback in planning agents. Second, we use a rule-based program instead of GPT-4 to obtain the ground truth plans based on the sampled tool sequences, which eliminates the possibility of hallucinated and incorrect plans. 


\subsection{Dataset quantity and quality}~\label{sec:data_stats}
\begin{table}[t]
\centering
\caption{The statistics of the \name dataset.}
\label{tab:dataset-stats}
 \resizebox{0.6\textwidth}{!}{
\begin{tabular}{lr}
\hline
Item                                     & Number         \\\hline \hline
Raw examples                             & 4427           \\
Human verified examples                  & 1565           \\
Human verified \& balanced examples      & 882            \\ 
- 1 / 2 / 3-tool examples                  & 70 / 159 / 653 \\  \hline
Tools                                    & 33             \\
- ML model / image processing / API        & 13 / 11 / 9   \\
Tool graphs                              & 317            \\
Avg. \# of unique queries per tool graph & 2.78          \\ \hline
\end{tabular}
}
\end{table}

Overall, \name contains a large quantity of \textbf{diverse ecologically-valid task} queries (see Figure~\ref{fig:dataset-examples}).
Each task is associated with \textbf{human-verified} and executable plans (Table~\ref{tab:dataset-stats}).
Concretely, there are a total of 4427 raw examples in \name, where 1565 have been verified to be correct by three human annotators. After additional filtering for a \textbf{balanced tool distribution} (See Appendix for more details), we select a subset of 882 examples for our evaluation.
Tasks are \textbf{granular in difficulty} with 70 queries that require a single tool, 159 need two tools, and 653 need three tools. 
In terms of tools, there are \textbf{33 unique tools in total across three different categories}, of which 13 are multi-modal machine learning models on HuggingFace, 11 are image processing modules from VisProg~\cite{gupta2022visual}, and 9 are free public APIs from RapidAPI\footnote{https://rapidapi.com/hub}. 
Our final dataset includes 317 representative tool graphs, where each graph has multiple queries. See more examples in the Appendix.

\section{Planning agent}
\begin{figure}[t]
    \centering
    \includegraphics[width=\textwidth]{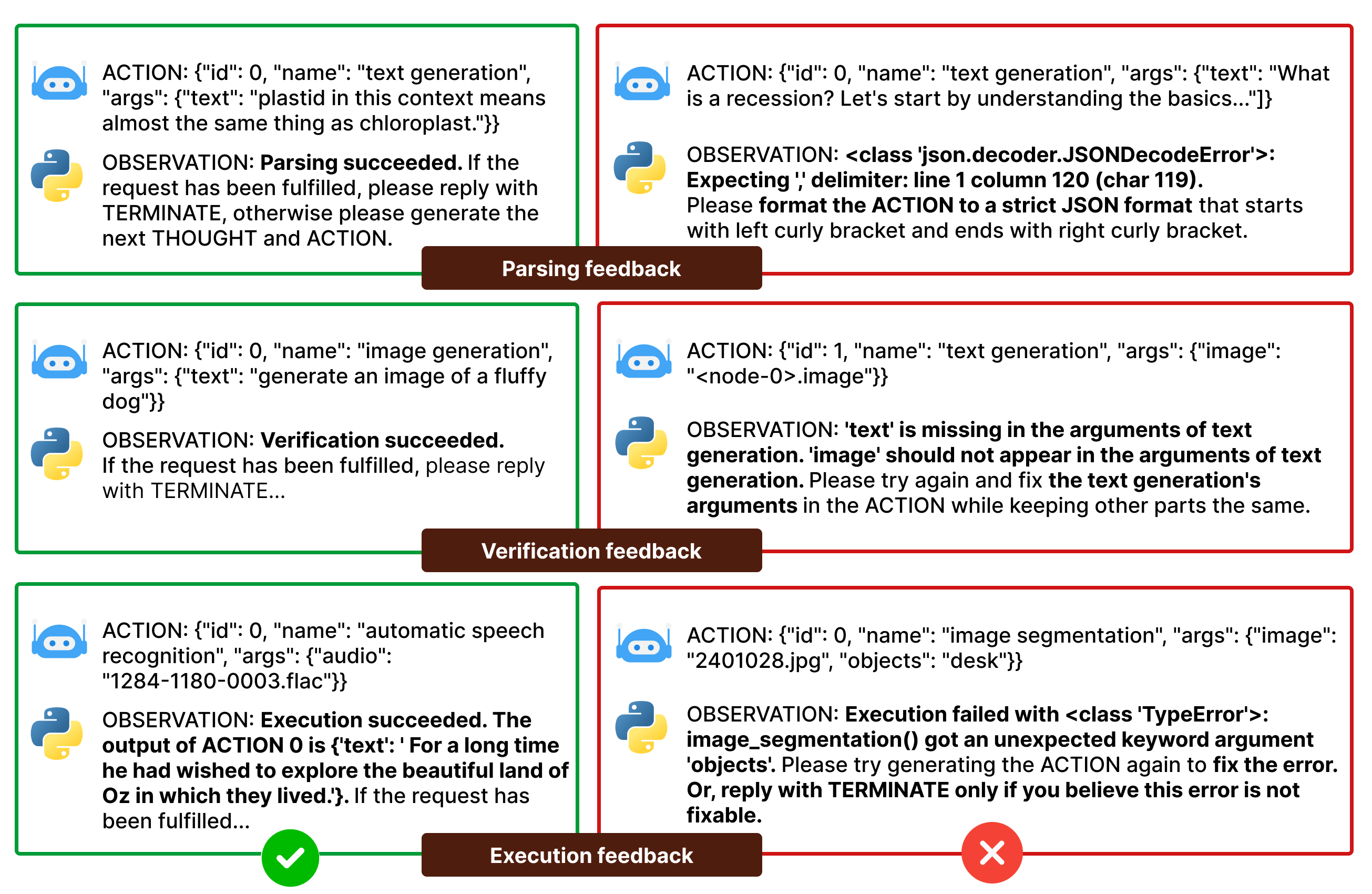}
    \caption{\textbf{Types of feedback.} We present examples of parsing, verification, and execution feedback in both success and failure cases.}
    \label{fig:feedback-types}
\end{figure}
To systematically evaluate the design space of planning agents, we design a modular planning system with these components: planning LLM, parser, verifier, and executor. We implement this system with AutoGen's framework~\cite{wu2023autogen}. Given the user query, the LLM must iteratively generate and refine the plan. Each iteration involves generating the whole or a part of the plan and receiving feedback on the generation. Given the raw text output from the LLM at the current iteration, the \name agent supports the following 3 kinds of feedback - \\

\noindent\textbf{Parsing feedback.} The parser attempts to parse the LLM text output to either JSON or code formats and returns an error message in case of parsing failures. \\

\noindent\textbf{Plan verification feedback.} The verifier checks the parsed output according to pre-defined rules and returns an error message in case of rule violations. Specifically, the verifier checks if the predicted tool exists in our provided tool list, if it forms a valid connection with the previous tool, and if the predicted argument names match the ones specified in the metadata document. \\

\noindent\textbf{Plan execution feedback.} In the case of JSON output, the executor calls the functions with specified arguments in a Python environment and returns the output or execution errors. In the case of code output, the code is directly executed with outputs or errors returned as feedback.

\section{Experiment}
Using our benchmark with a flexible agent design, we experiment with \modelnum instruction-tuned large language models of varying sizes (7 open-source and 3 proprietary) across different planning setups. We describe these evaluation setups in Sec.~\ref{sec:eval_setup}, metrics in Sec.~\ref{sec:metrics}, and our experimental findings in Sec.~\ref{sec:results}. 

\subsection{Setup}~\label{sec:eval_setup}
\begin{figure}[t]
    \centering
    \includegraphics[width=\textwidth]{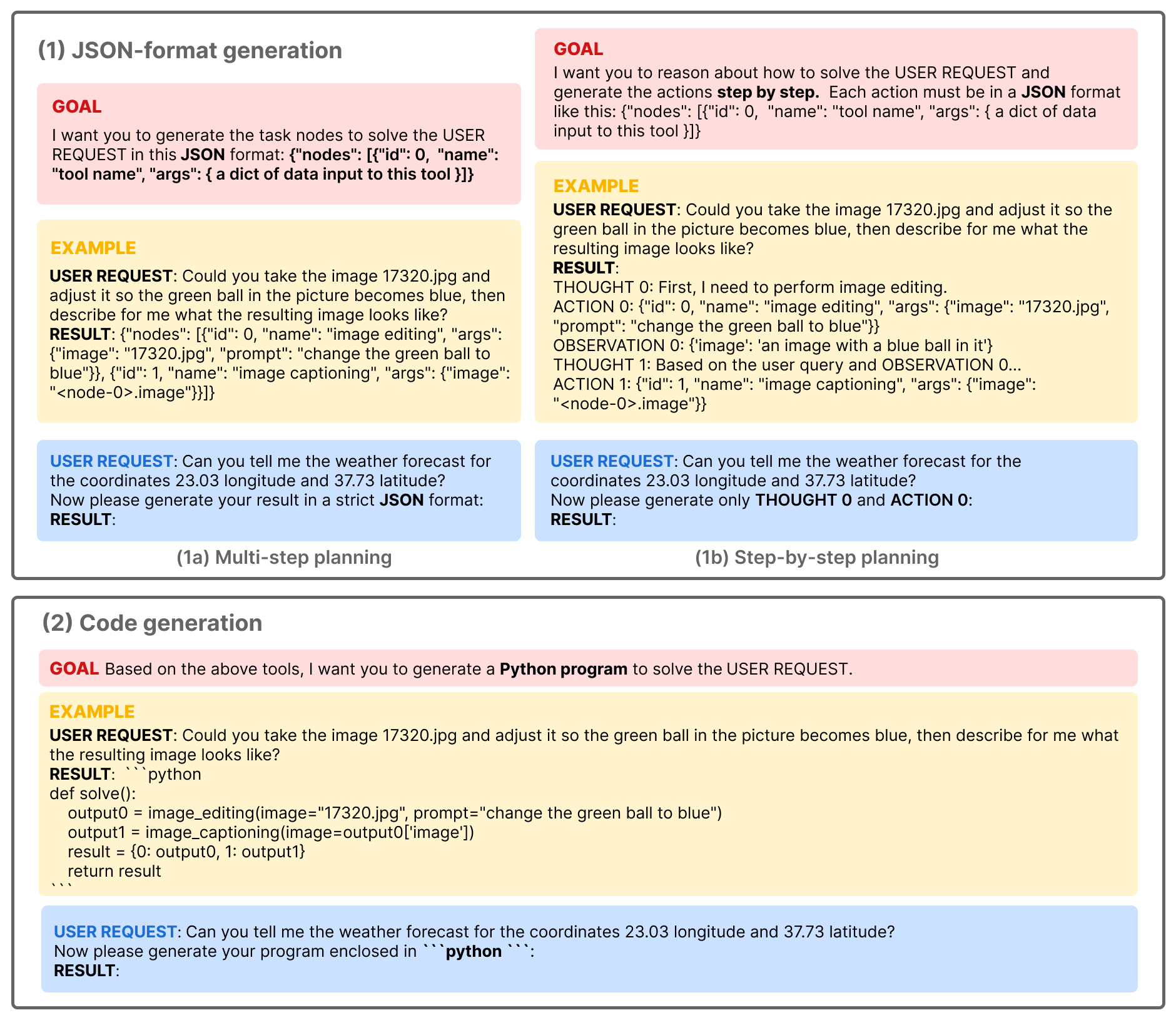}
    \caption{\textbf{Illustrating the three main planning setups in our evaluation:} (1a) multi-step and (1b) step-by-step JSON-format language generation \cite{yao2023react}, and (2) code generation. (See the Appendix for the full prompts). }
    \label{fig:plan-setups}
\end{figure}
We establish a unified framework to categorize LLMs' task planning setups along the three axes below. 
\noindent\textbf{Planning strategy:} Prior works formulate task planning as either step-by-step or multi-step planning\cite{yao2023react, qin2023toolllm, shen2023hugginggpt}. Step-by-step planning refers to the setup where a language model is instructed to predict only one action at a time (Figure \ref{fig:plan-setups} (1b)). On the other hand, in the setting of multi-step planning, a model can predict multiple actions at once (Figure \ref{fig:plan-setups} (1a)). 
\noindent\textbf{Plan format:} Additionally, existing works have also adopted different plan formats for tool use: often as code, pseudo-code, or predefined structured representations such as JSON \cite{shen2023hugginggpt, gupta2022visual, suris2023vipergpt}. In this work, we primarily focus on two of the common plan formats: JSON and code.  
\noindent\textbf{Feedback:} 
We experiment with three kinds of feedback - feedback from parsers, rule-based verifiers, and execution modules (Figure \ref{fig:feedback-types}). Nevertheless, our benchmark can be used to study other types of feedback, such as self-feedback \cite{madaan2023selfrefine}, which we leave to future work.

\subsection{Evaluation metrics}~\label{sec:metrics}
To holistically evaluate planning agents' performance on our benchmark, we adopt three main metrics: \textbf{tool-F1}, \textbf{argname-F1}, and \textbf{pass rate}. \textbf{Tool-F1} is defined as the F1 score of tool name prediction, where we treat each predicted tool name as one example and compare the set of predicted tool names to the groundtruth set of tools in each plan. Similarly, \textbf{argname-F1} is defined as the F1 score of argument name prediction for each tool, where we consider each (tool name, argument names) tuple as one example. Our implementation turns each tuple into a string and compares the set of predicted ``tool name-argument names'' strings to the labels. 
\textbf{Pass rate} is the percentage of predictions that execute successfully without any execution errors. It measures the executability but not the correctness of the predicted plans. We choose these three metrics because they assess two important aspects of planning and tool use: tool selection and tool invocation. A higher tool-F1 indicates better tool selection, whereas higher argname-F1 and pass rate imply improved tool invocation. 
To evaluate models with the same metrics in the code generation setup, we parse the generated code into an Abstract Syntax Tree (AST) with Python's AST module and extract the function names and argument names for calculating tool-F1 and argname-F1. 

We also provide argvalue-F1 in the Appendix for completeness but caution the reader about the challenges of evaluating argument values due to surface-form or syntactic differences in the values, particularly for free-form text arguments (e.g. the prompts in \tool{image generation} and \tool{text generation}). We report additional results on plan evaluation metrics, including overall plan accuracy, normalized edit distance, edge-F1, code-specific metrics such as AST accuracy and CodeBLEU, and plan execution accuracy in the Appendix.



\subsection{Results}~\label{sec:results}

\begin{figure}[t]
    \centering
    \includegraphics[width=\textwidth]{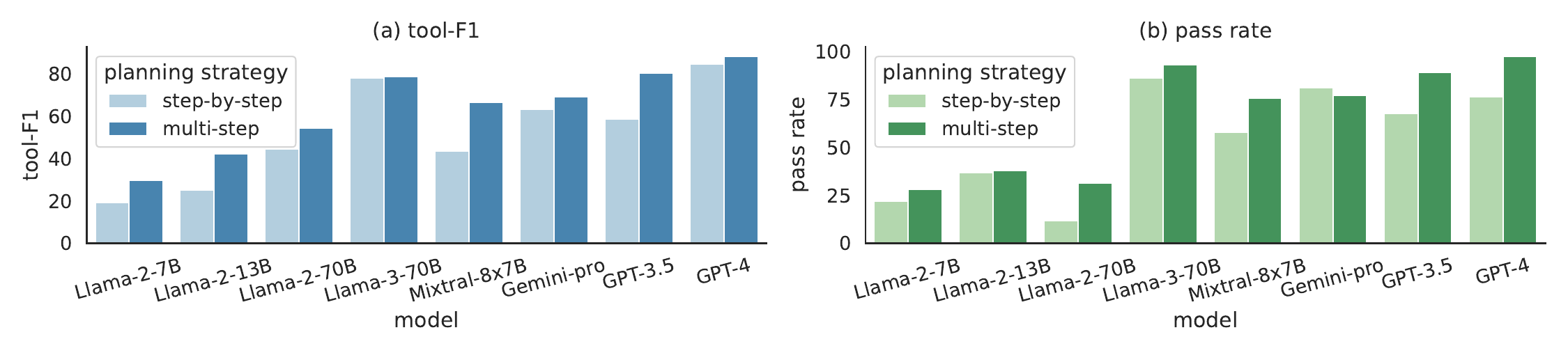}
    \caption{\textbf{Comparing planning strategies.} We find that models consistently perform better on tool-F1 under multi-step prediction compared to under step-by-step prediction regardless of their sizes. Similarly, all models except for Gemini-pro achieve a higher pass rate with multi-step prediction. }
    \label{fig:single-vs-multi}
\end{figure}

\begin{figure}
    \centering
    \includegraphics[width=0.9\textwidth]{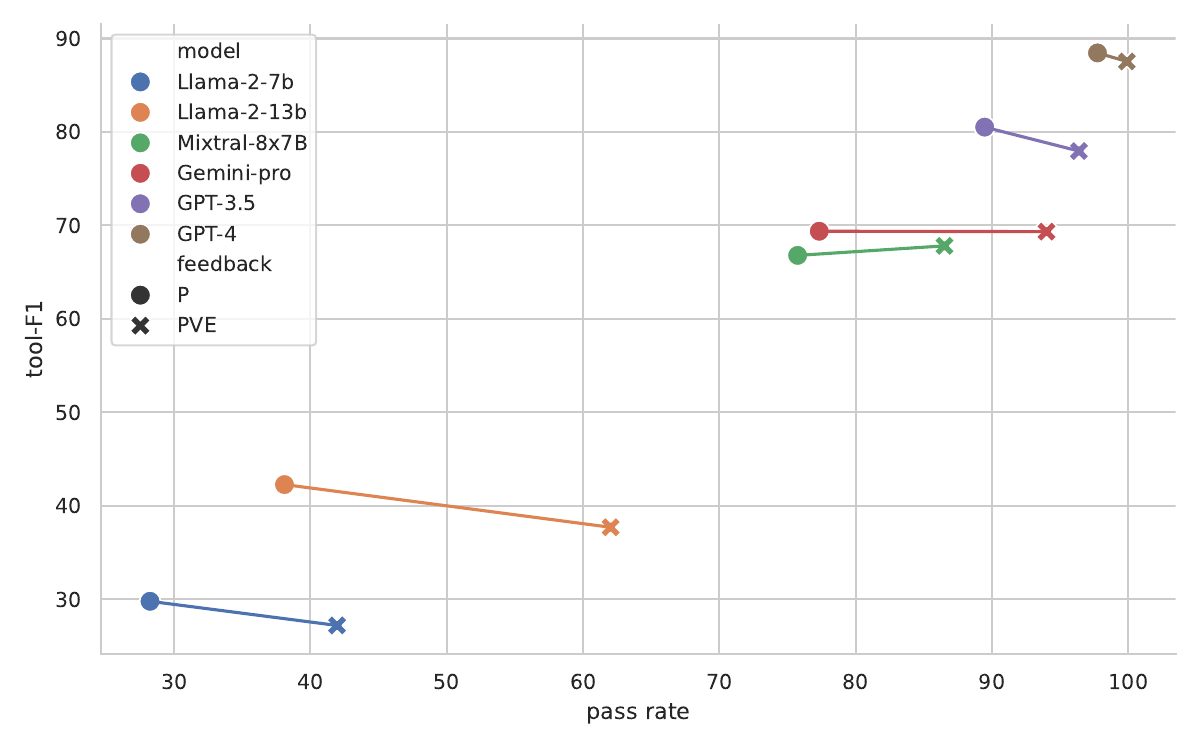}
    \caption{\textbf{Comparing tool-F1 and pass rate without vs. with feedback.} 
    We find that feedback greatly improves planning agents' pass rates across different model sizes, especially for Llama-7B/13B and Gemini-pro. However, feedback can also harm models' tool prediction performance and decrease their tool-F1 by a small amount (< 5\%). }
    \label{fig:tool_f1_vs_pass_rate}
\end{figure}

\begin{figure}[t]
    \centering
    \includegraphics[width=\textwidth]{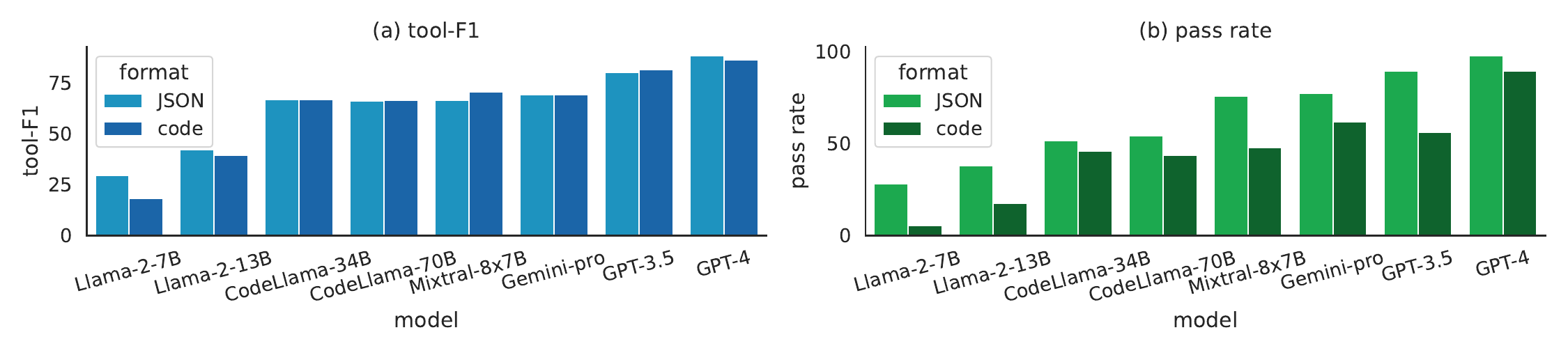}
    \caption{\textbf{Comparing plan formats.} We find that all models except for Llama-7-b perform comparably on tool-F1 with JSON-format and code generation. However, JSON-format generation leads to a much higher pass rate across all models.}
    \label{fig:dict-vs-code}
\end{figure}

We first highlight the key findings from our empirical analysis and then describe each finding in more detail: 

\begin{enumerate}
    \item All planning agents perform better on tool selection with multi-step planning than with step-by-step planning, with a smaller gap for larger models (Fig.~\ref{fig:single-vs-multi})
    \item Verification and execution feedback can help models improve tool invocation by predicting correct argument names and generating executable plans but can lead to worse tool selection due to wrong fixes (Tab.~\ref{tab:multi-dict-feedback} and Fig.~\ref{fig:tool_f1_vs_pass_rate})
    \item While models perform comparably on tool selection with JSON versus code generation, they produce more overall executable plans with JSON-format generation (Fig.~\ref{fig:dict-vs-code})
\end{enumerate}



\begingroup
\setlength{\tabcolsep}{5pt}
\begin{table}[t]
\caption{We present the tool-F1 and argname-F1 of models with various feedback, where P, V, and E represent parsing, verification, and execution feedback respectively. We use parsing feedback only (P) under multi-step planning and JSON-format language generation as the basis, while showing the $\Delta$ of those with other feedback combinations. We find that verification and execution feedback can improve models' performance on argument name prediction and pass rate, but can hurt tool selection.}
\label{tab:multi-dict-feedback}
\resizebox{\columnwidth}{!}{%
\begin{tabular}{lcccc|cccc|cccc}

                    & \multicolumn{4}{c|}{tool-F1}                                                                                            & \multicolumn{4}{c|}{argname-F1}     & \multicolumn{4}{c}{pass rate}                                                                   \\ \hline\hline
model               & \multicolumn{1}{c}{P} & \multicolumn{1}{c}{PV} & \multicolumn{1}{c}{PE} & \multicolumn{1}{c|}{PVE} & \multicolumn{1}{c}{P} & \multicolumn{1}{c}{PV} & \multicolumn{1}{c}{PE} & \multicolumn{1}{c|}{PVE}  & \multicolumn{1}{c}{P} & \multicolumn{1}{c}{PV} & \multicolumn{1}{c}{PE} & \multicolumn{1}{c}{PVE}           \\\hline
Llama-2-7B  & 29.78                  & \cellcolor{red!15}-2.94                      & \cellcolor{red!15}-2.59                       & \cellcolor{red!15}-2.58                              & 34.03                  & \cellcolor{green!15}2.03                       & \cellcolor{green!10}1.24                        & \cellcolor{green!10}1.15                               & 28.23     & \cellcolor{green!30}18.14 & \cellcolor{green!30}10.32 & \cellcolor{green!30}13.72 \\

Llama-2-13B & 42.27                  & \cellcolor{red!20}-3.45                      & \cellcolor{red!15}-2.78                       & \cellcolor{red!20}-4.57                              & 45.07                  & \cellcolor{green!20}3.94                       & \cellcolor{green!20}3.08                        & \cellcolor{green!20}3.29 & 38.10     & \cellcolor{green!40}29.93 & \cellcolor{green!50}32.99 & \cellcolor{green!40}23.92                               \\
Llama-2-70B & 54.40                  & \cellcolor{red!5}    -0.35	                  & \cellcolor{red!5}  -0.49	                    & \cellcolor{red!5}         -0.03                   &       52.49	            & \cellcolor{green!30}           12.87	            & \cellcolor{green!25}      8.97	                & \cellcolor{green!30} 12.60 &   31.52	   & \cellcolor{green!50} 39.80	& \cellcolor{green!40} 23.13	& \cellcolor{green!40}  29.59                            \\
Mixtral-8x7B        &  66.79 & \cellcolor{green!10}1.18   & \cellcolor{red!5}-0.11   & \cellcolor{red!5}-0.04          & 72.52 & \cellcolor{green!15}2.00   & \cellcolor{green!10}1.89    & \cellcolor{green!15}2.72           & 75.74     & \cellcolor{green!30}10.32 & \cellcolor{green!30}8.96  & \cellcolor{green!30}10.77    \\
Gemini-pro          & 69.38                  & \cellcolor{green!10}1.18                       & \cellcolor{red!5}-0.11                       & \cellcolor{red!5}-0.04                              & 73.37                  & \cellcolor{green!15}2.00                       & \cellcolor{green!10}1.89                        & \cellcolor{green!15}2.72               & 77.32     & \cellcolor{green!30}13.27 & \cellcolor{green!30}14.06 & \cellcolor{green!30}16.67                \\
Llama-3-70B & 78.73                 & \cellcolor{green!10}     1.54	                & \cellcolor{red!5}    -0.30	                  & \cellcolor{green!5}     0.70                         &        84.97	           & \cellcolor{green!5}    0.45	                   & \cellcolor{green!5}      -0.68	                  & \cellcolor{green!5} -0.08 &  92.40	  & \cellcolor{red!5} -0.45	 & \cellcolor{green!20} 4.31	& \cellcolor{green!20} 3.29                           \\
GPT-3.5-turbo-0125  & 80.52                  & \cellcolor{red!5}-0.65                      & \cellcolor{red!15}-2.80                       & \cellcolor{red!15}-2.56                              & 84.86                  & \cellcolor{green!5}0.65                       & \cellcolor{red!5}-0.92                       & \cellcolor{red!5}-0.86                    & 89.46     & \cellcolor{green!25}6.69  & \cellcolor{green!25}7.26  & \cellcolor{green!25}6.92          \\
GPT-4-0125-preview  & 88.46                  & \cellcolor{red!5}-0.60                      & \cellcolor{green!5}0.25                        & \cellcolor{red!5}-0.91                              & 89.81                  & \cellcolor{red!5}-0.18                      & \cellcolor{green!5}0.48                        & \cellcolor{green!5}0.32                    & 97.73     & \cellcolor{green!10}1.13  &   \cellcolor{red!10}-1.25    & \cellcolor{green!15}2.15            \\
GPT-4o-2024-05-13  &89.28	                  & \cellcolor{red!5}-0.22	                      & \cellcolor{green!5}0.48                       & \cellcolor{red!5}	-0.21	                              & 90.32	                 & \cellcolor{green!10}1.24	                       & \cellcolor{green!10}1.00	                        & \cellcolor{green!10}1.24                   &    96.37	  & \cellcolor{green!15} 2.61	 &   \cellcolor{red!5}  -0.45	  & \cellcolor{green!15}  2.15       \\

\hline \\
 \multicolumn{13}{l}{\small Note: we use the experiments with parsing feedback as the baseline to highlight external feedback's effects on tool selection} 
 \\
 \multicolumn{13}{l}{\small  and invocation instead of parsing. We include the results of experiments with no feedback in the Appendix. } \\
\end{tabular}
}
\end{table}
\endgroup
\begin{figure}
    \centering
    \includegraphics[width=0.8\textwidth]{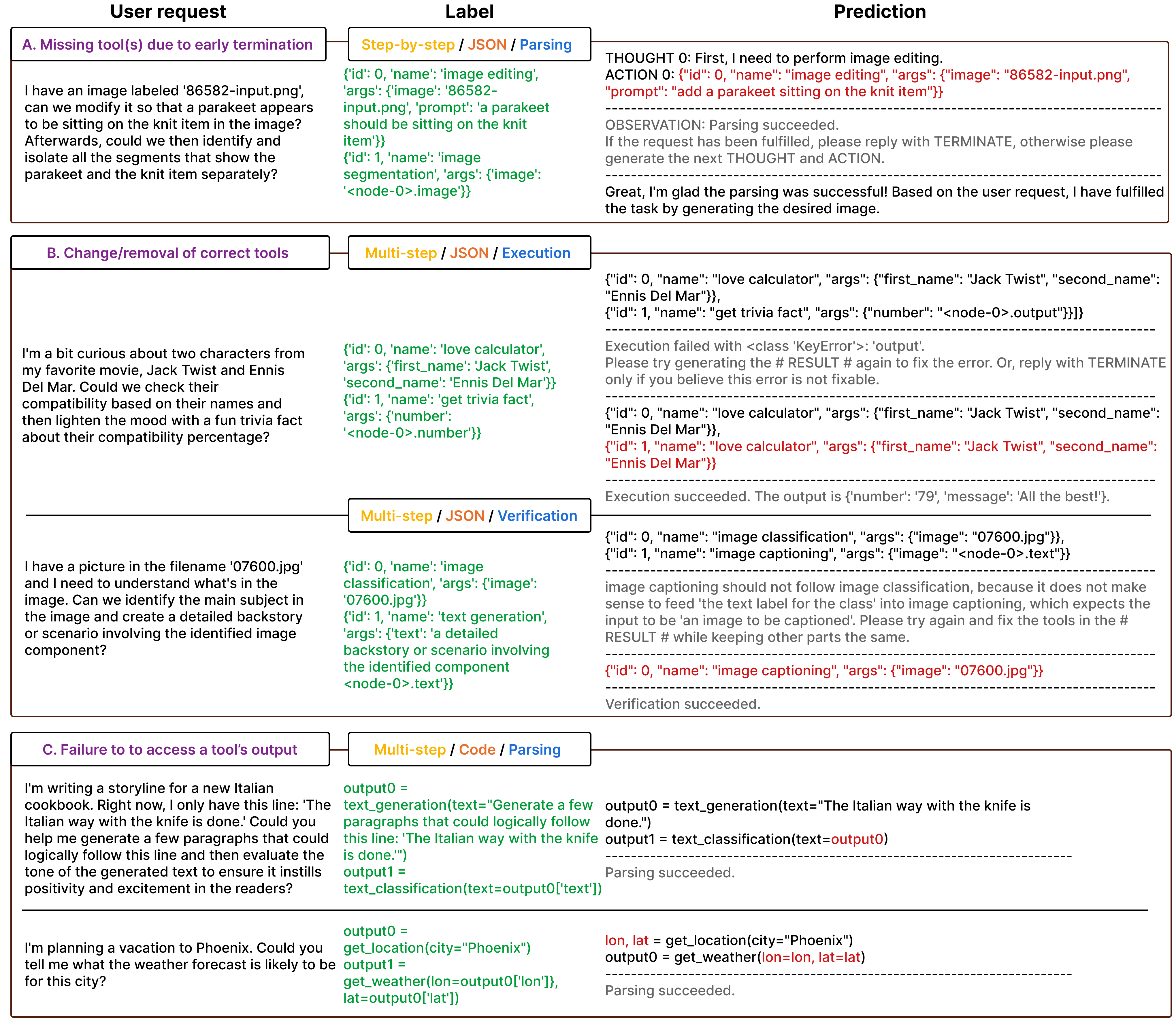}
    \caption{We present examples of three common errors (A-C) in different planning setups.}
    \label{fig:errors}
\end{figure}
\noindent\textbf{Models consistently perform better on tool-F1 and pass rate when instructed to perform multi-step planning instead of step-by-step planning.}
We find that all large language models achieve higher tool-F1 when they are instructed to perform multi-step planning compared to when they perform step-by-step prediction (Figure \ref{fig:single-vs-multi}), although the gap is smaller for more capable larger models like LLama-3-70B and GPT-4. All models except for Gemini-pro achieve a higher pass rate with multi-step planning. Among the \modelnum models we evaluated, Llama-2-7B, Llama-2-13B, and GPT-3.5 all showcase a large increase (>10\%) in performance with multi-step planning compared to step-by-step prediction, with the greatest increase of 21.8\% for GPT-3.5. Through qualitative analysis, we learn that when models are instructed to perform step-by-step prediction, they tend to ``Terminate'' after they receive positive feedback (e.g. ``verification/execution succeeded'') from the environment, disregarding whether the user request has been fulfilled. This means that they often predict fewer steps than required and miss necessary tools to resolve the requests (See Figure \ref{fig:errors} A). \\

\noindent\textbf{External feedback can improve planning agents' performance on argument name prediction and pass rate.}
We find that both verification and execution feedback can lead to slightly better argname-F1 and much higher pass rates (Table \ref{tab:multi-dict-feedback}), indicating that feedback can help models predict correct argument names and generate more executable plans. With feedback, most models can increase argname-F1 by around 1-4\% and pass rate by up to 20-30\% (Table \ref{tab:multi-dict-feedback}). There are only a few exceptions on GPT-3.5 and GPT-4, which already obtain relatively high performance without feedback  (Table \ref{tab:multi-dict-feedback}). 
Also, verification feedback can be more helpful than execution feedback on argument name prediction. In qualitative analysis, we find that this is because our verifier pinpoints where the error occurs and outputs a targeted and thus more helpful feedback message. On the other hand, the execution module returns the error message as it is, which can be vague and obscure, thus confusing the model and even resulting in wrong fixes (Figure \ref{fig:errors} B). 

While we see generally positive effects of feedback on argname-F1 and pass rate, we also observe that feedback can lead to a small decrease (< 5\%) in models' tool-F1. We observe that this is mainly because models can change some correct tools to the wrong ones or remove them even though the feedback instructs them to only fix the erroneous parts in the plan (Figure \ref{fig:errors} B). One way to mitigate this error can be using more fine-grained and localized feedback \cite{wu2023finegrained}. Additionally, neither verification feedback nor execution feedback provides useful information on the correctness of the tool selection that can increase tool-F1. Nevertheless, we also note that the decrease in tool-F1 with feedback is a lot smaller compared to the gains in pass rate (Figure \ref{fig:tool_f1_vs_pass_rate}), which suggests feedback can greatly improve tool invocation at a small cost to tool selection. \\


\noindent\textbf{Models perform comparably on tool-F1 with JSON-format and code generation but much worse on pass rate with code generation.} We learn that plan formats can also influence models' tool use performance (Figure \ref{fig:dict-vs-code}), especially on the executability of the generated plans. Concretely, our experiments show that while all models except for Llama-2-7B achieve similar tool-F1s (<3\% difference) with JSON-format generation and code generation, they all suffer from a large drop in pass rate with code generation. Upon qualitative analysis, we find that one common execution error in code generation is failing to access the output from a tool (See Figure \ref{fig:errors} C).
While the same error also happens to JSON-format generation, it occurs less frequently due to the more rigid structure of JSON. These results suggest that JSON-format generation is preferable to code generation when the executability of generated plans matters.


\section{Discussion}
\subsection{Limitations}
There are a few limitations to our work. 
First, \name only considers sequential task plans, which represent a majority of real-world user requests. However, some tasks might require dynamic task plans depending on the output for one subtask~\cite{grunde2023designing}. Dynamic plans require a more complex tool graph sampling procedure. 
Second, as our main goal is to study the effects of planning formulations and feedback, we do not investigate another dimension of planning design: prompt style. We use direct and ReACT-style~\cite{yao2023react} prompting and exclude more sophisticated prompting strategies such as tree-of-thoughts prompting~\cite{yao2023tree, wang2023selfconsistency}. 
Third, as some tools in our benchmark are suboptimal, generative and/or non-deterministic, we only conducted evaluation of the execution results on a limited subset (See Appendix). Finally, we have only evaluated LLM planners because of their advanced abilities and leave the evaluation of multi-modal planners to future work. 

\subsection{Conclusion}
In conclusion, we highlight three major contributions of our work: first, we introduce a new benchmark \name to support comprehensive and rigorous evaluation of tool-use abilities of planning agents for multi-step multi-modal tasks. \name contains a large and diverse set of queries and human-verified and executable plans; second, we characterize the design space of existing tool-use methods and conducted a systematic study of \modelnum\ LLMs with different design choices, including planning formulations, plan formats and various types of feedback; finally, our experiments reveal three takeaways, suggesting that current generation of LLMs demonstrate gains in tool-planning performance on \name when using multi-step planning, outputting plans in JSON format, and using feedback. We hope \name enables further investigation into better planning formulations that incorporate richer and more diverse kinds of feedback for solving multi-step, multi-modal tasks.  
\section*{Acknowledgement}
This work was partially funded by a Sony grant. It was also made possible because of OpenAI’s credit grant. We also thank Zeyu Tang for his help with figures and annotations, and Jiafei Duan, Chenhao Zheng, and Dylan Bunarto for their help with data annotations.

\clearpage
\appendix
\section{Additional data}
\begin{figure}
    \centering
\includegraphics[width=\textwidth]{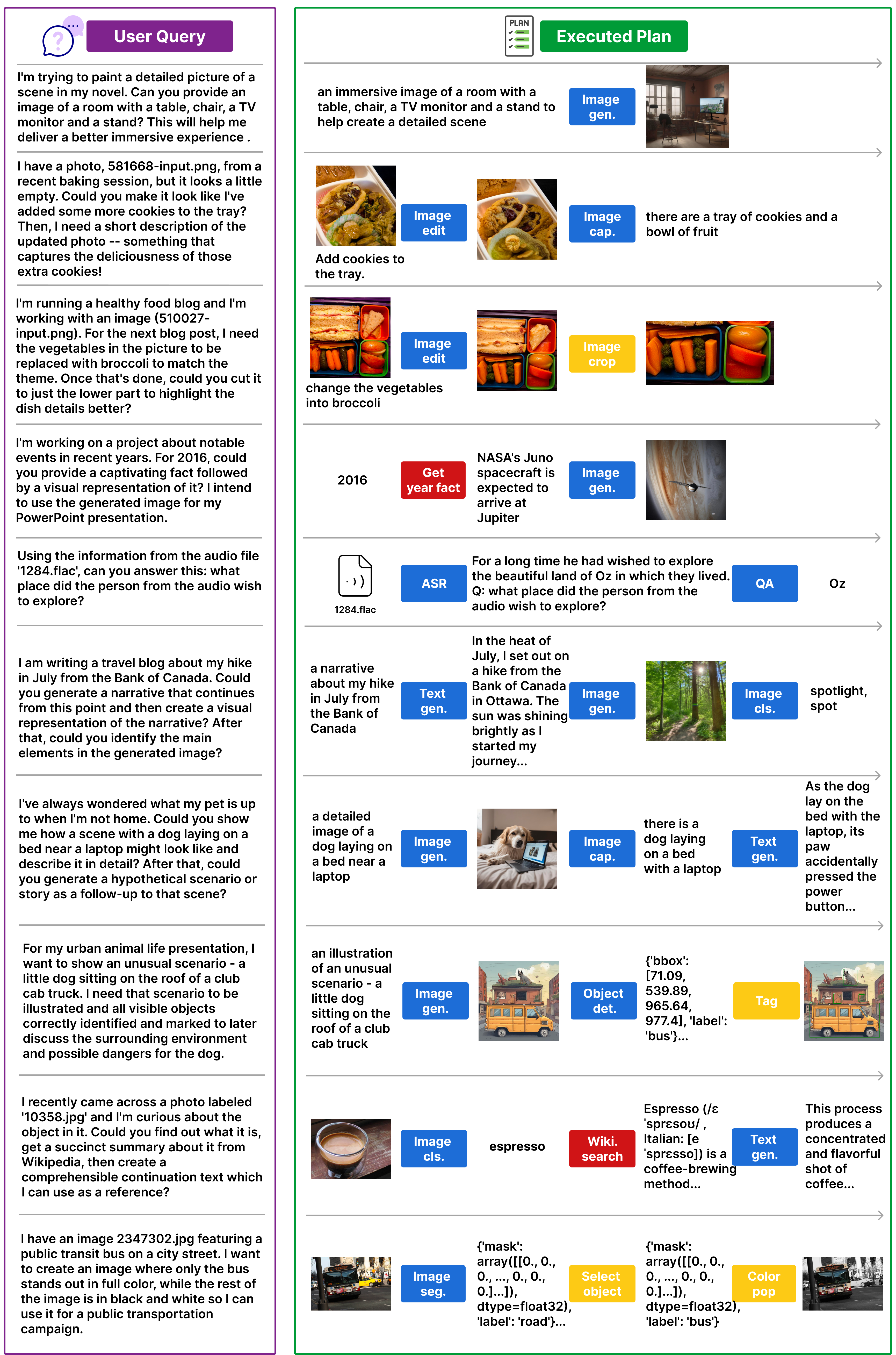}
    \caption{We present additional examples of query-plan pairs along with the execution results of the plans in \name.}
    \label{fig:add-dataset-examples}
\end{figure}
\begingroup
\setlength{\tabcolsep}{5pt}
\begin{table}[h]
\centering
\caption{We list all 33 tools across three categories - ML models, public APIs, and image processing modules - in \name.}
\label{tab:all-tools}
\resizebox{\columnwidth}{!}{%
\begin{tabular}{ll}
Tool category    & Tool name   \\\hline\hline
ML model         & text generation, text summarization, text classification, question answering, \\
   & optical character recognition, image generation, image editing, image \\
   & captioning, image classification, image segmentation, object detection, visual \\
   & question answering, automatic speech recognition \\
Public APIs      & get weather, get location, get math fact, get trivia fact, get year fact, get date
\\
   & fact, search movie, love calculator, wikipedia simple search        \\
Image processing & image crop, image crop top, image crop bottom, image crop left, image crop 
\\
   & right, select object, count, tag, color pop, emoji, background blur            \\\hline                                                                                                                                           
\end{tabular}
}
\end{table}
\endgroup
We present more examples of query-plan pairs of \name in Figure \ref{fig:add-dataset-examples}, and a complete list of all 33 tools in Table \ref{tab:all-tools}. For more details about the tools and their implementation, please refer to our Github codebase\footnote{https://github.com/RAIVNLab/mnms}. 

\section{Dataset generation}
\begin{figure}
    \centering
    \includegraphics[width=\textwidth]{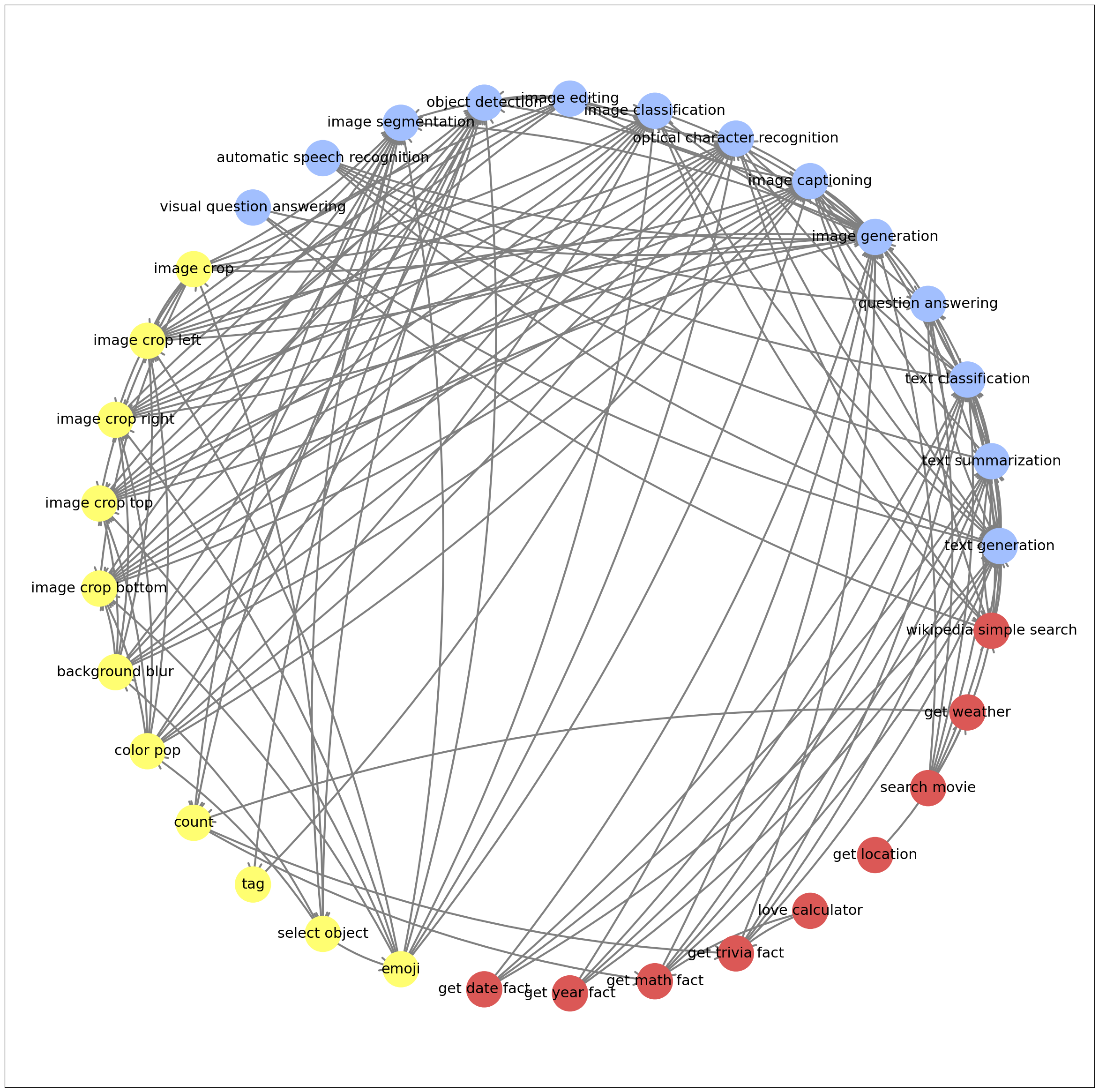}
    \caption{The full tool graph consists of 33 unique tools as nodes (red = public APIs, yellow = image processing tools, blue = machine learning models) and valid connections between them as edges. }
    \label{fig:tool-graph}
\end{figure}
\begin{figure}[t]
    \centering
    \includegraphics[width=\textwidth]{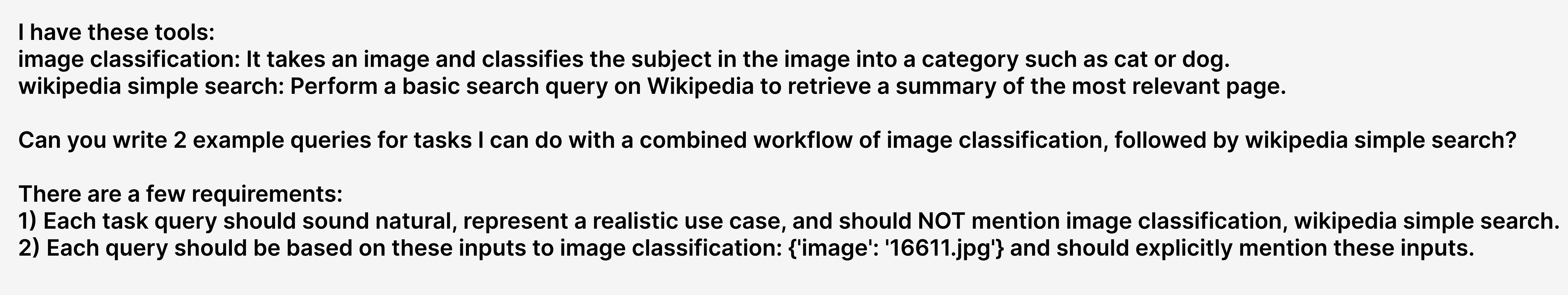}
    \caption{\textbf{Query generation prompt.} We present the full prompt used for query generation.}
    \label{fig:query-gen-prompt}
\end{figure}
\begin{figure}[t]
    \centering
    \includegraphics[width=\textwidth]{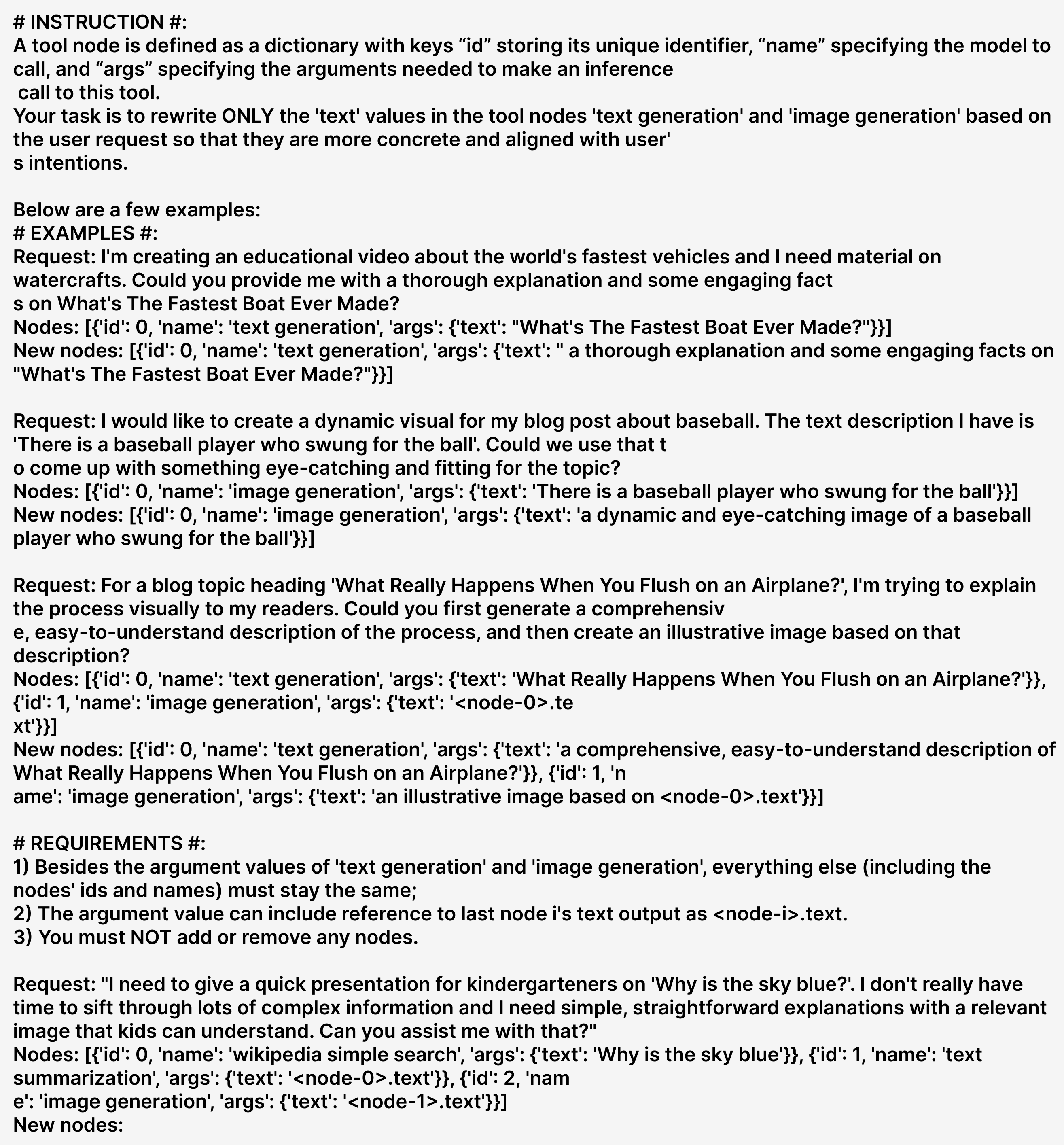}
    \caption{\textbf{Argument value rewrite prompt.} We present the full prompt used for rewriting the argument values of \tool{text generation} and \tool{image generation}.}
    \label{fig:argvalue-rewrite-prompt}
\end{figure}
\begin{figure}[t]
    \centering
    \includegraphics[width=0.8\textwidth]{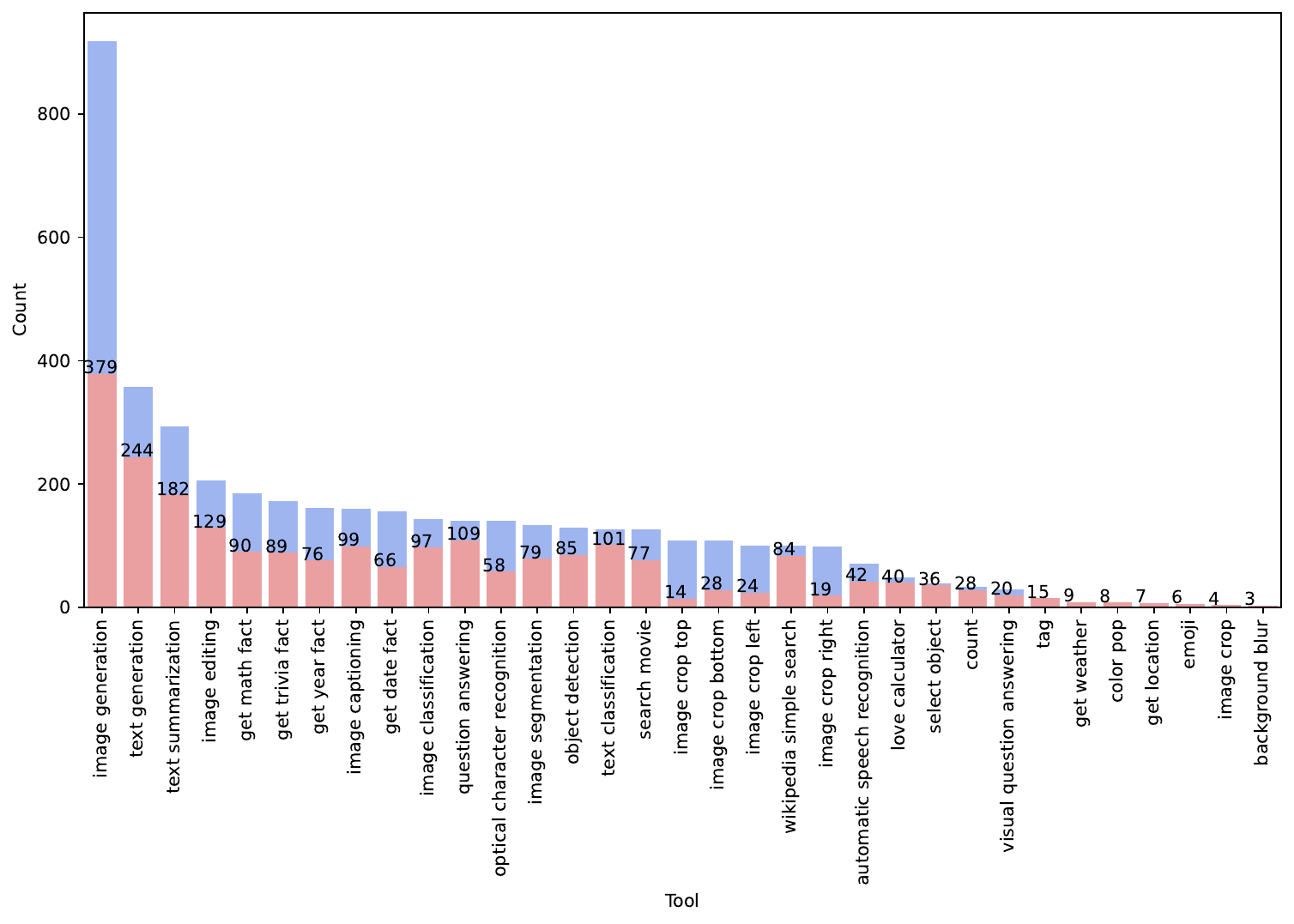}
    \caption{\textbf{Tool distribution before and after filtering.} }
    \label{fig:tool-freq}
\end{figure}
\subsection{Tool graph}
We include a visualization of the full tool graph used in our dataset generation pipeline (Figure \ref{fig:tool-graph}).  

\subsection{Prompts} We generate the queries with the prompt in Figure \ref{fig:query-gen-prompt}, and rewrite the argument values of \tool{text generation} and \tool{image generation} with the prompt shown in Figure \ref{fig:argvalue-rewrite-prompt}.

\subsection{Human verification statistics}
The pairwise agreement rates among the 3 annotators are 74.95\%, 81.43\%, 70.88\%, and the average pairwise agreement rate is 75.75\% (std=4.34\%). 

\subsection{Data filtering} We perform two types of data filtering on the 1565 human-verified examples: (1) we manually filter out 349 examples with poor execution results, especially those where intermediate tools return wrong or empty outputs (e.g. when \tool{question answering} is the second tool in the sequence and outputs an empty string); (2) we filter out a total of 334 examples whose plans involve \tool{image generation} and have more than 4 unique queries. We perform the second filtering step because of two reasons. First, the frequency of the tools initially follows the distribution in Figure \ref{fig:tool-freq} (blue), where \tool{image generation} has a much higher count -- 918 -- than other tools. Thus, we would like to reduce the frequency of \tool{image generation} in the dataset while maintaining the frequency of rare tools. To achieve this while also preserving the diversity of tool plans, we choose to filter out examples whose plans have 5-10 unique queries, as the average number of unique requests per tool plan before filtering is 4.20. We end up filtering out 40\% (or 349) of these examples. After these two filtering steps, we are left with 882 examples in total that follow the distribution in Figure \ref{fig:tool-freq} (red).

\subsection{Alternative plans} In addition to the one human verified groundtruth plan, we have also generated alternative plans to supplement our evaluation. Concretely, we generate these alternative plans in three steps: first, we generate a set of syntactically valid (i.e. the alternative tool's input and output types are correct) and semantically valid (i.e. the alternative tool performs the same functionality as the original tool) alternative tools for each tool in our toolset; second, we manually verify their validity and only keep the human-verified valid tools in the alternative tools set; finally, we compose all valid tools at each position in the plan to obtain all combinations as the total set of valid plans. To generate the syntactically valid tools, we create a graph with both data (including input and output) and tools as nodes, and we obtain the syntactic alternative tools $t_o^{alt}$ of the original tool $t_o$ by searching for all possible paths from $t_o$'s input to its output. As for semantic alternative tools, we prompt GPT-4 to generate these for each tool in the toolset. For example, for the plan \tool{image classification} $\rightarrow$ \tool{text generation}, we first obtain alternative tools to each of them. 
For \tool{image classification}, its syntactic alternative tools include \tool{image captioning} and \tool{visual question answering} as these tools' inputs both include one image and their outputs are a text -- the same as \tool{image classification}'s. In addition, GPT-4 identifies \tool{object detection} as a semantic alternative to \tool{image classification}. On the other hand, there are no human-verified alternative tools to \tool{text generation}. Therefore, there are a total of 3 alternative plans to \tool{image classification} $\rightarrow$ \tool{text generation}. 

\section{Planning agent}
\begin{figure}[t]
    \centering
    \includegraphics[width=\textwidth]{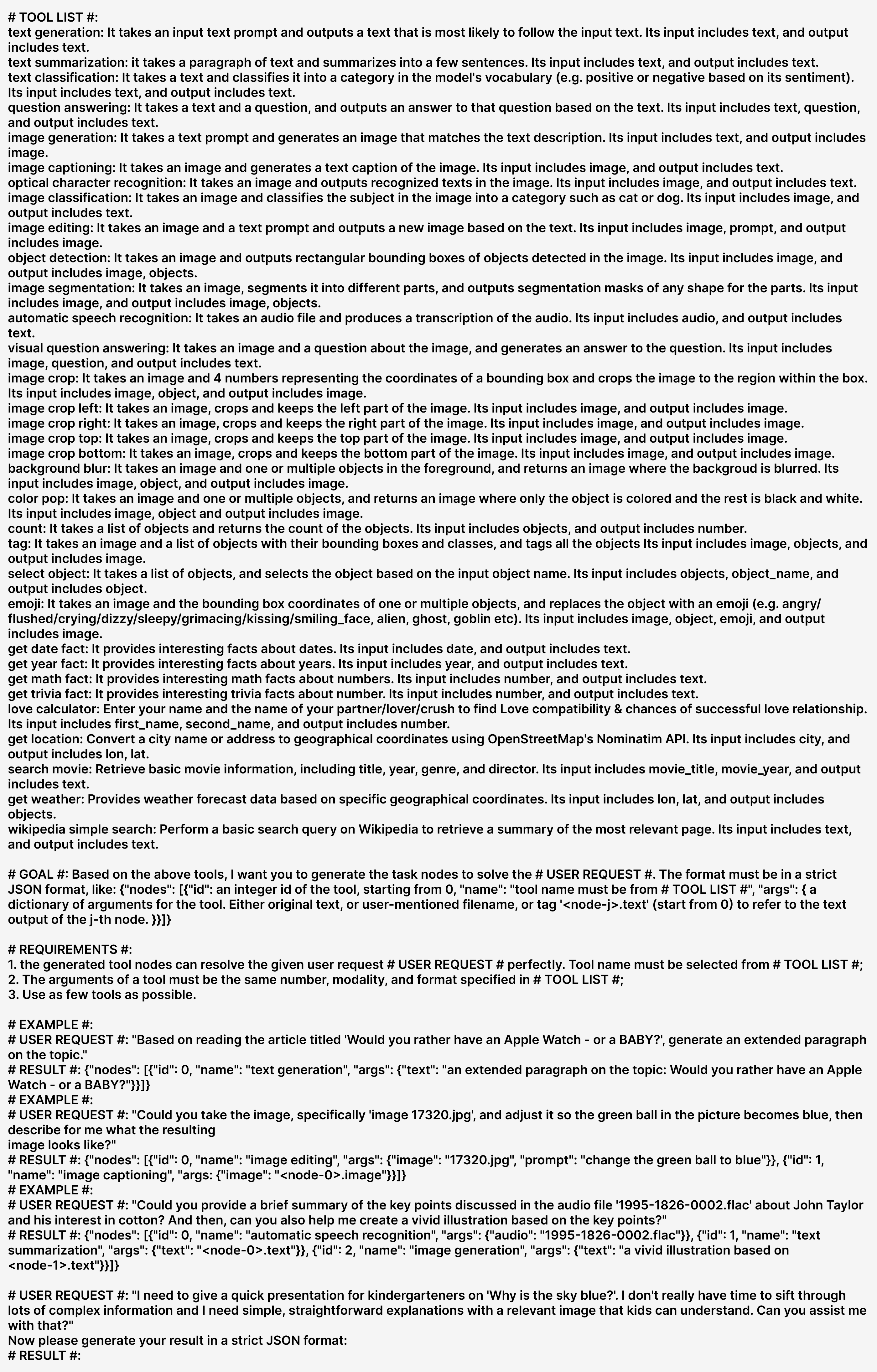}
    \caption{\textbf{Multi-step planning prompt.} We present the full prompt used for multi-step planning.}
    \label{fig:multi-step-plan-prompt}
\end{figure}
\begin{figure}[t]
    \centering
    \includegraphics[width=\textwidth]{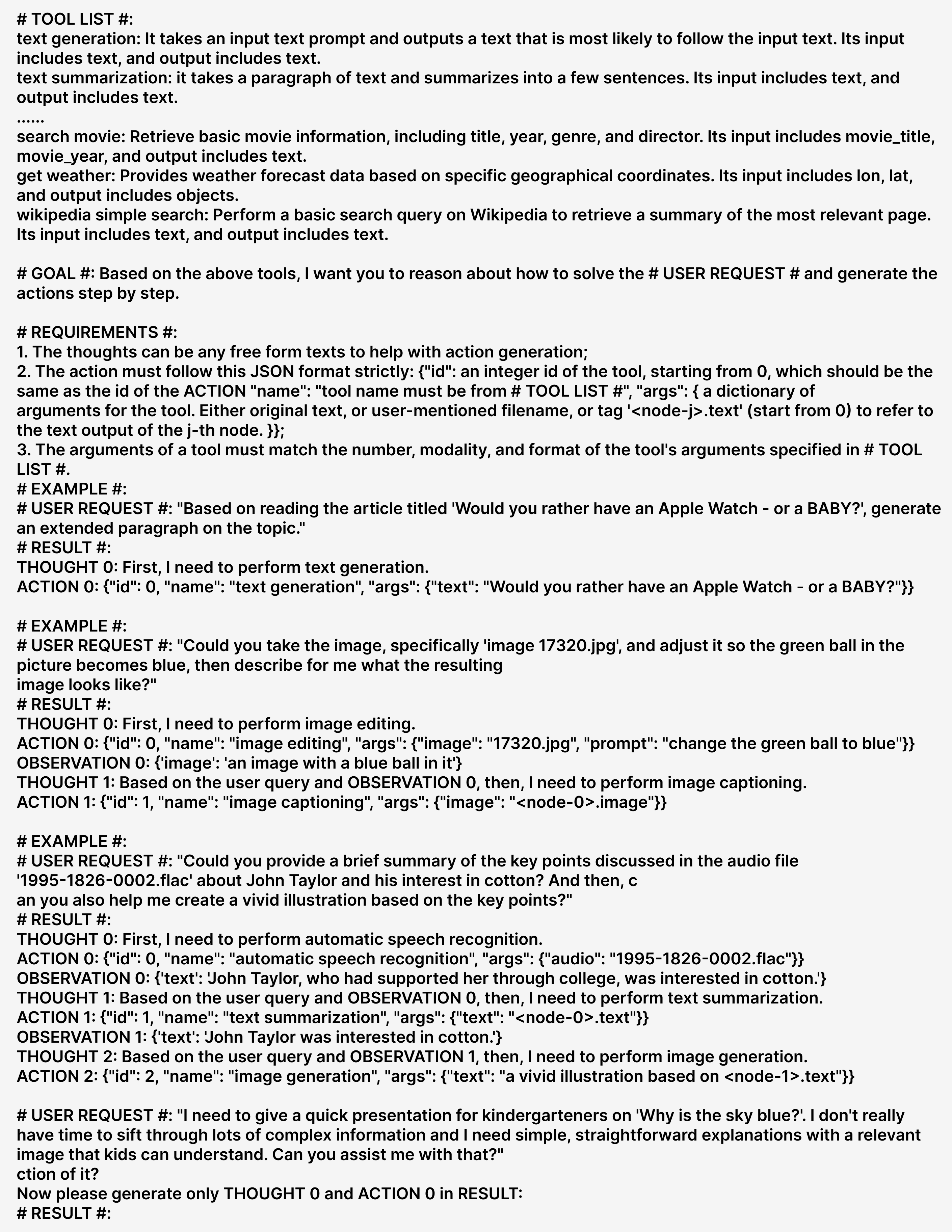}
    \caption{\textbf{Step-by-step planning prompt.} We present the full prompt used for step-by-step planning.}
    \label{fig:step-by-step-plan-prompt}
\end{figure}
\begin{figure}[t]
    \centering
    \includegraphics[width=\textwidth]{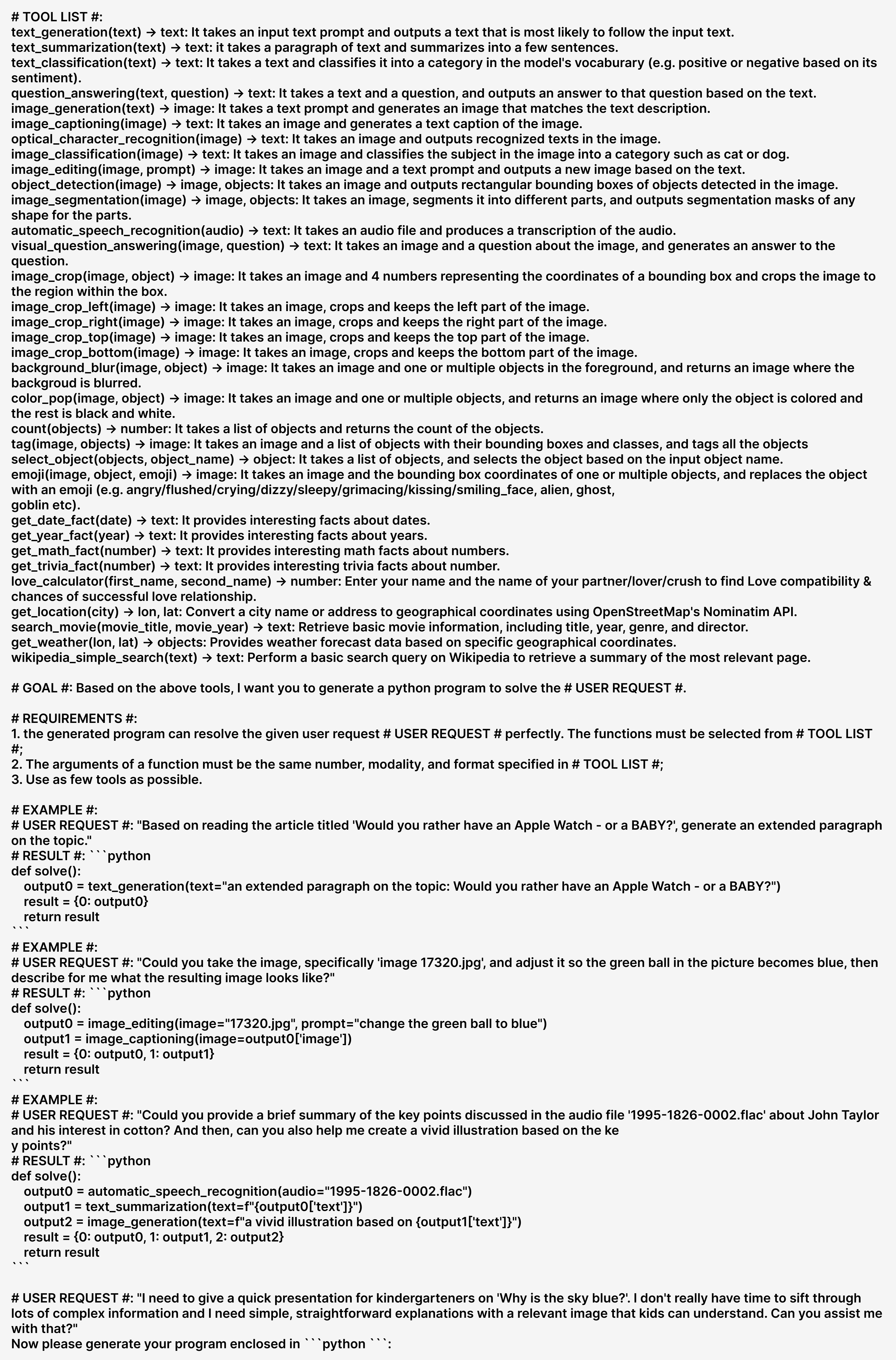}
    \caption{\textbf{Code generation prompt.} We present the full prompt used for code generation.}
    \label{fig:code-gen-prompt}
\end{figure}
We present the full prompts used for multi-step JSON-format planning (Figure \ref{fig:multi-step-plan-prompt}), step-by-step JSON-format planning (Figure \ref{fig:step-by-step-plan-prompt}, excluding details in the TOOL LIST which are the same as the ones in Figure \ref{fig:multi-step-plan-prompt}) as well as code generation (Figure \ref{fig:code-gen-prompt}).

\section{Additional plan evaluation results}
\begingroup
\setlength{\tabcolsep}{5pt}
\begin{table}[t]
\caption{We present the tool-F1, argname-F1 and pass rate of models with various feedback, where P, V, and E represent parsing, verification, and execution feedback respectively. We use no feedback only (N/A) under multi-step planning and JSON-format language generation as the basis, while showing the $\Delta$ of those with other feedback combinations compared to no feedback.}
\label{tab:no-feedback}
\resizebox{\columnwidth}{!}{%
\begin{tabular}{lccccc|ccccc|ccccc}
                   & \multicolumn{5}{c|}{tool-F1}           & \multicolumn{5}{c|}{argname-F1}       & \multicolumn{5}{c}{pass rate}        \\ \hline\hline
model              & N/A   & P     & PV    & PE    & PVE   & N/A   & P     & PV    & PE    & PVE  & N/A   & P    & PV    & PE    & PVE   \\\hline
Llama-2-7b         & 27.37 & \cellcolor{green!15}2.41  & \cellcolor{red!5}-0.53 & \cellcolor{red!5}-0.18 & \cellcolor{red!5}-0.18 & 30.71 & \cellcolor{green!20}3.31  & \cellcolor{green!20}5.34  & \cellcolor{green!20}4.56  & \cellcolor{green!20}4.47 & 24.83 & \cellcolor{green!20}3.40 & \cellcolor{green!40}21.54 & \cellcolor{green!30}13.72 & \cellcolor{green!30}17.12 \\
Llama-2-13b        & 40.30 & \cellcolor{green!10}1.97  & \cellcolor{red!10}-1.48 & \cellcolor{red!5}-0.80 & \cellcolor{red!15}-2.60 & 43.30 & \cellcolor{green!10}1.77  & \cellcolor{green!20}5.72  & \cellcolor{green!20}4.86  & \cellcolor{green!20}5.06 & 37.30 & \cellcolor{green!5}0.79 & \cellcolor{green!50}30.73 & \cellcolor{green!50}33.79 & \cellcolor{green!40}24.72 \\
Mixtral-8x7B       & 65.06 & \cellcolor{green!10}1.73  & \cellcolor{green!5}0.88  & \cellcolor{green!5}0.15  & \cellcolor{green!20}2.75  & 73.00 & \cellcolor{red!5}-0.49 & \cellcolor{green!10}1.12  & \cellcolor{red!5}-0.14 & \cellcolor{green!5}0.85 & 69.61 & \cellcolor{green!25}6.12 & \cellcolor{green!30}16.44 & \cellcolor{green!30}15.08 & \cellcolor{green!30}16.89 \\
Gemini-pro         & 68.57 & \cellcolor{green!5}0.80  & \cellcolor{green!10}1.98  & \cellcolor{green!5}0.69  & \cellcolor{green!5}0.76  & 72.79 & \cellcolor{green!5}0.58  & \cellcolor{green!15}2.58  & \cellcolor{green!15}2.47  & \cellcolor{green!20}3.30 & 73.92 & \cellcolor{green!20}3.40 & \cellcolor{green!30}16.67 & \cellcolor{green!30}17.46 & \cellcolor{green!40}20.07 \\
GPT-3.5-turbo-0125 & 79.83 & \cellcolor{green!5}0.68  & \cellcolor{green!5}0.03  & \cellcolor{red!15}-2.11 & \cellcolor{red!10}-1.88 & 83.94 & \cellcolor{green!5}0.92  & \cellcolor{green!10}1.57  & \cellcolor{green!5}0.00  & \cellcolor{green!5}0.06 & 88.44 & \cellcolor{green!10}1.02 & \cellcolor{green!25}7.71  & \cellcolor{green!25}8.28  & \cellcolor{green!25}7.94  \\
GPT-4-0125-preview & 88.96 & \cellcolor{red!5}-0.50 & \cellcolor{red!10}-1.10 & \cellcolor{red!5}-0.26 & \cellcolor{red!10}-1.42 & 89.88 & \cellcolor{red!5}-0.07 & \cellcolor{red!5}-0.25 & \cellcolor{green!5}0.41  & \cellcolor{green!5}0.25 & 97.39 & \cellcolor{green!5}0.34 & \cellcolor{green!10}1.47  & \cellcolor{red!5}-0.91 & \cellcolor{green!15}2.49 \\\hline
\end{tabular}
}
\end{table}
\endgroup
\begingroup
\setlength{\tabcolsep}{10pt}
\begin{table}[t]
\centering
\caption{\textbf{argvalue-F1.} We present the argvalue-F1 of step-by-step and multi-step planning with JSON-format generation and different types of feedback.}
\label{tab:argvalue-f1}
\resizebox{0.8\columnwidth}{!}{%
\begin{tabular}{llrrrr}
                                    &              & \multicolumn{4}{l}{argvalue-F1} \\ \hline\hline
model                               & strategy     & P      & PV     & PE    & PVE   \\ \hline
\multirow{2}{*}{Llama-2-7b}         & step-by-step & 4.63   & 8.28   & 9.68  & 9.57  \\
                                    & multi-step   & 10.34  & 9.88   & 9.47  & 10.57 \\
\multirow{2}{*}{Llama-2-13b}        & step-by-step & 7.10   & 11.30  & 12.59 & 12.64 \\
                                    & multi-step   & 15.39  & 17.11  & 15.84 & 16.71 \\
\multirow{2}{*}{Mixtral-8x7B}       & step-by-step & 20.44  & 24.32  &   21.77	  &    21.69   \\
                                    & multi-step   & 36.45  & 36.70  & 35.70 & 36.73 \\
\multirow{2}{*}{Gemini-pro}         & step-by-step & 32.28  & 27.81  & 32.22 & 31.37 \\
                                    & multi-step   & 37.22  & 39.89  & 36.30 & 38.33 \\
\multirow{2}{*}{GPT-3.5-turbo-0125} & step-by-step & 29.58  & 28.32  & 23.61 & 23.24 \\
                                    & multi-step   & 45.64  & 46.54  & 45.15 & 45.56 \\
\multirow{2}{*}{GPT-4-0125-preview} & step-by-step & 47.37  & 46.91  & 34.49 & 34.84 \\
                                    & multi-step   & 51.02  & 51.08  & 51.70 & 51.99 \\ \hline
\end{tabular}
}
\end{table}
\endgroup
\begingroup
\setlength{\tabcolsep}{10pt}
\begin{table}[t]
\centering
\caption{\textbf{edge-F1.} We present the edge-F1 of step-by-step and multi-step planning with JSON-format generation and different types of feedback.}
\label{tab:edge-f1}
\resizebox{0.8\columnwidth}{!}{%
\begin{tabular}{llrrrr}
                                    &              & \multicolumn{4}{l}{edge-F1}   \\ \hline \hline
model                               & strategy     & P     & PV    & PE    & PVE   \\\hline
\multirow{2}{*}{Llama-2-7b}         & step-by-step & 1.61  & 2.35  & 3.98  & 3.37  \\
                                    & multi-step   & 12.44 & 11.61 & 12.10 & 11.27 \\
\multirow{2}{*}{Llama-2-13b}        & step-by-step & 5.74  & 6.22  & 6.96  & 8.22  \\
                                    & multi-step   & 23.27 & 23.98 & 24.00 & 23.58 \\
\multirow{2}{*}{Mixtral-8x7B}       & step-by-step & 15.41 & 21.88 &   24.00    &     24.77  \\
                                    & multi-step   & 55.72 & 53.10 & 53.08 & 53.52 \\
\multirow{2}{*}{Gemini-pro}         & step-by-step & 41.39 & 17.86 & 45.82 & 45.08 \\
                                    & multi-step   & 54.98 & 56.63 & 53.60 & 55.22 \\
\multirow{2}{*}{GPT-3.5-turbo-0125} & step-by-step & 31.37 & 27.23 & 39.40 & 39.72 \\
                                    & multi-step   & 69.52 & 71.03 & 67.98 & 69.05 \\
\multirow{2}{*}{GPT-4-0125-preview} & step-by-step & 73.68 & 72.67 & 68.28 & 68.12 \\
                                    & multi-step   & 78.80 & 78.79 & 79.47 & 79.60 \\ \hline
\end{tabular}
}
\end{table}
\endgroup
\begingroup
\setlength{\tabcolsep}{10pt}
\begin{table}[t]
\centering
\caption{\textbf{Normalized edit distance.} We present the normalized edit distance of step-by-step and multi-step planning with JSON-format generation and different types of feedback.}
\label{tab:norm_edit_dist}
\resizebox{0.8\columnwidth}{!}{%
\begin{tabular}{llllll}
                                    &              & \multicolumn{4}{l}{Normalized edit distance $\downarrow$} \\ \hline\hline
model                               & strategy     & P      & PV     & PE    & PVE   \\ \hline
\multirow{2}{*}{Llama-2-7b}         & step-by-step & 80.39     & 75.24     & 76.00     & 74.55    \\
                                    & multi-step   & 61.14     & 64.43     & 62.82     & 63.12    \\
\multirow{2}{*}{Llama-2-13b}        & step-by-step & 72.81     & 68.57     & 68.60     & 67.84    \\
                                    & multi-step   & 47.57     & 48.69     & 49.63     & 49.73    \\
\multirow{2}{*}{Mixtral-8x7B}       & step-by-step & 60.81     & 56.28     &     56.86    &     56.78     \\
                                    & multi-step   & 23.97     & 25.97     & 26.64     & 26.26    \\
\multirow{2}{*}{Gemini-pro}         & step-by-step & 36.23     & 47.89     & 34.70     & 36.00    \\
                                    & multi-step   & 28.18     & 27.34     & 25.96     & 24.77    \\
\multirow{2}{*}{GPT-3.5-turbo-0125} & step-by-step & 51.46     & 52.38     & 47.93     & 47.44    \\
                                    & multi-step   & 16.08     & 15.55     & 17.44     & 17.86    \\
\multirow{2}{*}{GPT-4-0125-preview} & step-by-step & 14.26     & 14.70     & 16.92     & 16.62    \\
                                    & multi-step   & 10.96     & 11.39     & 10.59     & 10.81   
                                    
                                     \\ \hline
\end{tabular}
}
\end{table}
\endgroup
\begingroup
\setlength{\tabcolsep}{5pt}
\begin{table}[t]
\caption{\textbf{Plan accuracy}}
\label{tab:plan-acc}
\resizebox{\columnwidth}{!}{%
\begin{tabular}{llrrrr|rrrr}
                                              \multicolumn{2}{r}{Plan accuracy}  & \multicolumn{4}{c}{(tool)} & \multicolumn{4}{|c}{(tool+argname)} \\ \hline\hline
model                               & strategy     & P        & PV       & PE       & PVE     & P          & PV         & PE         & PVE       \\\hline
\multirow{2}{*}{Llama-2-7b}         & step-by-step & 1.13     & 2.27     & 3.29     & 3.29    & 1.13       & 2.27       & 3.29       & 3.29      \\
                                    & multi-step   & 4.20     & 3.40     & 2.95     & 4.20    & 2.95       & 3.29       & 2.04       & 3.51      \\
\multirow{2}{*}{Llama-2-13b}        & step-by-step & 1.25     & 3.17     & 3.74     & 4.99    & 1.13       & 3.17       & 3.74       & 4.99      \\
                                    & multi-step   & 11.90    & 13.83    & 10.88    & 12.13   & 9.52       & 13.27      & 9.98       & 11.79     \\
\multirow{2}{*}{Mixtral-8x7B}       & step-by-step & 9.41     & 14.63    &   14.06	       &   14.97      & 9.41       & 14.63      &   14.06	         &   14.97       \\
                                    & multi-step   & 45.80    & 45.12    & 45.12    & 45.35   & 45.12      & 45.01      & 44.90      & 45.24     \\
\multirow{2}{*}{Gemini-pro}         & step-by-step & 24.83    & 10.66    & 30.27    & 28.57   & 24.38      & 10.66      & 30.16      & 28.57     \\
                                    & multi-step   & 41.84    & 42.18    & 40.70    & 42.40   & 40.48      & 42.18      & 40.59      & 42.40     \\
\multirow{2}{*}{GPT-3.5-turbo-0125} & step-by-step & 19.27    & 14.97    & 18.59    & 19.16   & 19.27      & 14.97      & 18.59      & 19.16     \\
                                    & multi-step   & 59.64    & 60.20    & 57.48    & 58.39   & 59.52      & 60.20      & 57.48      & 58.39     \\
\multirow{2}{*}{GPT-4-0125-preview} & step-by-step & 61.68    & 60.88    & 51.93    & 53.17   & 61.68      & 60.88      & 51.93      & 53.17     \\
                                    & multi-step   & 70.63    & 69.50    & 71.43    & 70.63   & 70.63      & 69.50      & 71.43      & 70.63 \\\hline    
\end{tabular}
}
\end{table}
\endgroup
\begingroup
\setlength{\tabcolsep}{5pt}
\begin{table}[t]
\caption{\textbf{$\Delta$ in plan accuracy considering alternative plans.}}
\label{tab:plan-acc-alt}
\resizebox{\columnwidth}{!}{%
\begin{tabular}{llrrrr|rrrr}
                                              \multicolumn{2}{r}{ $\Delta$ in plan accuracy}  & \multicolumn{4}{c}{(tool)} & \multicolumn{4}{|c}{(tool+argname)} \\ \hline\hline
model                               & strategy     & P        & PV       & PE       & PVE     & P          & PV         & PE         & PVE       \\\hline
\multirow{2}{*}{Llama-2-7b}         & step-by-step & 0.00     & \cellcolor{green!5}0.11     & \cellcolor{green!5}0.11     & \cellcolor{green!5}0.11    & 0.00       & \cellcolor{green!5}0.11       & \cellcolor{green!5}0.11       & \cellcolor{green!5}0.11      \\
                                    & multi-step   & \cellcolor{green!5}0.79     & \cellcolor{green!5}0.34     & \cellcolor{green!5}0.68     & \cellcolor{green!5}0.57    & 0.00       & \cellcolor{green!5}0.11       & \cellcolor{green!5}0.11       & \cellcolor{green!5}0.23      \\
\multirow{2}{*}{Llama-2-13b}        & step-by-step & \cellcolor{green!5}0.57     & \cellcolor{green!5}0.57     & \cellcolor{green!5}0.68     & \cellcolor{green!5}0.91    & \cellcolor{green!5}0.45       & \cellcolor{green!5}0.57       & \cellcolor{green!5}0.68       & \cellcolor{green!5}0.91      \\
                                    & multi-step   & \cellcolor{green!10}1.36     & \cellcolor{green!10}1.47     & \cellcolor{green!10}1.47     & \cellcolor{green!10}1.47    & \cellcolor{green!5}0.91       & \cellcolor{green!10}1.36       & \cellcolor{green!10}1.25       & \cellcolor{green!10}1.25      \\
\multirow{2}{*}{Mixtral-8x7B}       & step-by-step & \cellcolor{green!5}0.79     & \cellcolor{green!15}2.15     &     \cellcolor{green!10}1.93     &    \cellcolor{green!15}2.04       & \cellcolor{green!5}0.79       & \cellcolor{green!10}1.93       &      \cellcolor{green!10}1.93 	      &    \cellcolor{green!10}1.93      \\
                                    & multi-step   & \cellcolor{green!20}4.08     & \cellcolor{green!20}3.40     & \cellcolor{green!20}3.74     & \cellcolor{green!15}2.83    & \cellcolor{green!20}3.40       & \cellcolor{green!20}3.40       & \cellcolor{green!20}3.29       & \cellcolor{green!15}2.61      \\
\multirow{2}{*}{Gemini-pro}         & step-by-step & \cellcolor{green!10}1.36     & \cellcolor{green!15}2.83     & \cellcolor{green!15}2.49     & \cellcolor{green!10}1.93    & \cellcolor{green!10}1.36       & \cellcolor{green!15}2.83       & \cellcolor{green!15}2.38       & \cellcolor{green!10}1.93      \\
                                    & multi-step   & \cellcolor{green!20}3.74     & \cellcolor{green!15}2.83     & \cellcolor{green!20}4.65     & \cellcolor{green!20}3.51    & \cellcolor{green!20}3.40       & \cellcolor{green!15}2.83       & \cellcolor{green!20}4.65       & \cellcolor{green!20}3.51      \\
\multirow{2}{*}{GPT-3.5-turbo-0125} & step-by-step & \cellcolor{green!10}1.02     & \cellcolor{green!5}0.34     & \cellcolor{green!10}1.02     & \cellcolor{green!5}0.68    & \cellcolor{green!10}1.02       &\cellcolor{green!5} 0.34       & \cellcolor{green!10}1.02       & \cellcolor{green!5}0.68      \\
                                    & multi-step   & \cellcolor{green!20}3.17     & \cellcolor{green!20}3.06     & \cellcolor{green!20}3.40     & \cellcolor{green!20}3.74    & \cellcolor{green!20}3.17       & \cellcolor{green!20}3.06       & \cellcolor{green!20}3.40       & \cellcolor{green!20}3.74      \\
\multirow{2}{*}{GPT-4-0125-preview} & step-by-step & \cellcolor{green!15}2.15     & \cellcolor{green!10}1.81     & \cellcolor{green!15}2.95     & \cellcolor{green!20}3.06    & \cellcolor{green!15}2.15       & \cellcolor{green!10}1.81       & \cellcolor{green!15}2.95       & \cellcolor{green!20}3.06      \\
                                    & multi-step   & \cellcolor{green!10}1.81     & \cellcolor{green!10}1.81     & \cellcolor{green!10}1.59     & \cellcolor{green!10}1.59    & \cellcolor{green!10}1.81       & \cellcolor{green!10}1.81       & \cellcolor{green!10}1.59       & \cellcolor{green!10}1.59      \\\hline    
\end{tabular}
}
\end{table}
\endgroup
\begingroup
\setlength{\tabcolsep}{5pt}
\begin{table}[t]
\caption{\textbf{Plan accuracy considering argument values}}
\label{tab:plan-acc-argvalue}
\resizebox{\columnwidth}{!}{%
\begin{tabular}{llrrrr|rrrr}
                                              \multicolumn{2}{l}{Plan accuracy (tool+argname+argvalue)}  & \multicolumn{4}{c}{exact matching} & \multicolumn{4}{|c}{entailment} \\ \hline\hline
model                               & strategy     & P        & PV       & PE       & PVE     & P          & PV         & PE         & PVE       \\\hline
\multirow{2}{*}{Llama-2-7b}         & step-by-step & 0.57         & 1.02         & 1.81         & 1.59         & 0.91             & 1.81             & 2.95             & 2.38             \\
                                    & multi-step   & 0.57         & 0.34         & 0.23         & 0.57         & 1.02             & 1.59             & 0.68             & 1.59             \\
\multirow{2}{*}{Llama-2-13b}        & step-by-step & 0.57         & 1.70         & 2.04         & 2.27         & 0.91             & 2.49             & 2.83             & 3.51             \\
                                    & multi-step   & 2.04         & 2.72         & 2.38         & 2.49         & 5.44             & 7.48             & 5.78             & 6.24             \\
\multirow{2}{*}{Mixtral-8x7B}       & step-by-step & 2.72         & 5.44         &    3.51	          &   3.51           & 6.12             & 9.86             &  7.03	                &       7.37           \\
                                    & multi-step   & 9.75         & 10.09        & 9.52         & 10.77        & 28.00            & 29.14            & 28.68            & 29.48            \\
\multirow{2}{*}{Gemini-pro}         & step-by-step & 7.03         & 5.78         & 7.48         & 6.58         & 15.42            & 9.52             & 17.12            & 15.19            \\
                                    & multi-step   & 8.39         & 11.34        & 9.07         & 11.45        & 24.15            & 27.89            & 24.83            & 27.66            \\
\multirow{2}{*}{GPT-3.5-turbo-0125} & step-by-step & 6.46         & 5.33         & 2.38         & 2.72         & 12.93            & 10.20            & 7.14             & 8.05             \\
                                    & multi-step   & 13.61        & 14.29        & 13.61        & 14.06        & 34.81            & 36.85            & 34.92            & 35.83            \\
\multirow{2}{*}{GPT-4-0125-preview} & step-by-step & 11.68        & 11.00        & 6.35         & 6.24         & 34.35            & 32.65            & 19.73            & 20.29            \\
                                    & multi-step   & 14.85        & 14.97        & 15.19        & 15.53        & 41.04            & 40.70            & 43.20            & 42.97   \\\hline    
\end{tabular}
}
\end{table}
\endgroup
\begingroup
\setlength{\tabcolsep}{5pt}
\begin{table}[t]
\caption{\textbf{Plan accuracy by number of tools with \textit{multi-step} planning}}
\label{tab:plan-acc-breakdown}
\resizebox{\columnwidth}{!}{%
\begin{tabular}{llrrrr|rrrr}
\multicolumn{2}{c}{Plan   accuracy}               & \multicolumn{4}{c}{(tool)}    & \multicolumn{4}{|c}{(tool + argname)} \\ \hline\hline
model                               & \# of tools & P     & PV    & PE    & PVE   & P       & PV      & PE      & PVE    \\ \hline
\multirow{3}{*}{Llama-2-7b}         & 1           & 1.43  & 4.29  & 1.43  & 1.43  & 1.43    & 4.29    & 1.43    & 1.43   \\
                                    & 2           & 10.06 & 9.43  & 6.92  & 9.43  & 8.18    & 9.43    & 5.66    & 7.55   \\
                                    & 3           & 3.06  & 1.84  & 2.14  & 3.22  & 1.84    & 1.68    & 1.23    & 2.76   \\ \hline
\multirow{3}{*}{Llama-2-13b}        & 1           & 7.14  & 18.57 & 10.00 & 8.57  & 7.14    & 17.14   & 10.00   & 8.57   \\
                                    & 2           & 18.87 & 23.90 & 16.98 & 18.24 & 16.98   & 23.27   & 15.09   & 17.61  \\
                                    & 3           & 10.72 & 10.87 & 9.49  & 11.03 & 7.96    & 10.41   & 8.73    & 10.72  \\ \hline
\multirow{3}{*}{Mixtral-8x7B}       & 1           & 70.00 & 71.43 & 71.43 & 71.43 & 68.57   & 71.43   & 71.43   & 71.43  \\
                                    & 2           & 55.97 & 55.97 & 55.97 & 57.86 & 55.97   & 55.35   & 55.97   & 57.23  \\
                                    & 3           & 40.74 & 39.66 & 39.66 & 39.51 & 39.97   & 39.66   & 39.36   & 39.51  \\ \hline
\multirow{3}{*}{Gemini-pro}         & 1           & 65.71 & 74.29 & 74.29 & 78.57 & 65.71   & 74.29   & 74.29   & 78.57  \\
                                    & 2           & 49.06 & 50.31 & 49.69 & 52.83 & 48.43   & 50.31   & 49.06   & 52.83  \\
                                    & 3           & 37.52 & 36.75 & 34.92 & 35.99 & 35.83   & 36.75   & 34.92   & 35.99  \\ \hline
\multirow{3}{*}{GPT-3.5-turbo-0125} & 1           & 74.29 & 75.71 & 71.43 & 71.43 & 74.29   & 75.71   & 71.43   & 71.43  \\
                                    & 2           & 72.96 & 72.33 & 69.81 & 70.44 & 72.96   & 72.33   & 69.81   & 70.44  \\
                                    & 3           & 54.82 & 55.59 & 52.99 & 54.06 & 54.67   & 55.59   & 52.99   & 54.06  \\ \hline
\multirow{3}{*}{GPT-4-0125-preview} & 1           & 78.57 & 78.57 & 81.43 & 81.43 & 78.57   & 78.57   & 81.43   & 81.43  \\
                                    & 2           & 80.50 & 79.25 & 78.62 & 78.62 & 80.50   & 79.25   & 78.62   & 78.62  \\
                                    & 3           & 67.38 & 66.16 & 68.61 & 67.53 & 67.38   & 66.16   & 68.61   & 67.53 \\ \hline
\end{tabular}
}
\end{table}
\endgroup
\begingroup
\setlength{\tabcolsep}{5pt}
\begin{table}[t]
\caption{\textbf{Plan accuracy by number of tools with \textit{step-by-step} planning}}
\label{tab:plan-acc-breakdown-local}
\resizebox{\columnwidth}{!}{%
\begin{tabular}{llrrrr|rrrr}
\multicolumn{2}{c}{Plan   accuracy}               & \multicolumn{4}{c}{(tool)}    & \multicolumn{4}{|c}{(tool + argname)} \\ \hline\hline
model                               & \# of tools & P     & PV    & PE    & PVE   & P       & PV      & PE      & PVE    \\ \hline
\multirow{3}{*}{Llama-2-7b}         & 1 & 14.29 & 24.29 & 34.29 & 32.86 & 14.29 & 24.29 & 34.29 & 32.86 \\
                                    & 2 & 0.00  & 0.63  & 2.52  & 3.14  & 0.00  & 0.63  & 2.52  & 3.14  \\
                                    & 3 & 0.00  & 0.31  & 0.15  & 0.15  & 0.00  & 0.31  & 0.15  & 0.15  \\ \hline
\multirow{3}{*}{Llama-2-13b}        & 1 & 11.43 & 32.86 & 31.43 & 35.71 & 10.00 & 32.86 & 31.43 & 35.71 \\
                                    & 2 & 1.89  & 0.63  & 5.03  & 6.29  & 1.89  & 0.63  & 5.03  & 6.29  \\
                                    & 3 & 0.00  & 0.61  & 0.46  & 1.38  & 0.00  & 0.61  & 0.46  & 1.38  \\ \hline
\multirow{3}{*}{Mixtral-8x7B}       & 1 & 40.00 & 67.14 & 51.43 & 52.86 & 40.00 & 67.14 & 51.43 & 52.86 \\
                                    & 2 & 22.64 & 30.19 & 28.93 & 32.70 & 22.64 & 30.19 & 28.93 & 32.70 \\
                                    & 3 & 2.91  & 5.21  & 6.43  & 6.58  & 2.91  & 5.21  & 6.43  & 6.58  \\ \hline
\multirow{3}{*}{Gemini-pro}         & 1 & 52.86 & 78.57 & 62.86 & 55.71 & 52.86 & 78.57 & 62.86 & 55.71 \\
                                    & 2 & 36.48 & 10.69 & 42.77 & 42.14 & 35.85 & 10.69 & 42.77 & 42.14 \\
                                    & 3 & 18.99 & 5.97  & 23.74 & 22.36 & 18.53 & 5.97  & 23.58 & 22.36 \\ \hline
\multirow{3}{*}{GPT-3.5-turbo-0125} & 1 & 67.14 & 72.86 & 32.86 & 31.43 & 67.14 & 72.86 & 32.86 & 31.43 \\
                                    & 2 & 28.93 & 18.87 & 37.74 & 37.74 & 28.93 & 18.87 & 37.74 & 37.74 \\
                                    & 3 & 11.79 & 7.81  & 12.40 & 13.32 & 11.79 & 7.81  & 12.40 & 13.32 \\ \hline
\multirow{3}{*}{GPT-4-0125-preview} & 1 & 84.29 & 82.86 & 80.00 & 84.29 & 84.29 & 82.86 & 80.00 & 84.29 \\
                                    & 2 & 70.44 & 70.44 & 71.70 & 72.96 & 70.44 & 70.44 & 71.70 & 72.96 \\
                                    & 3 & 57.12 & 56.20 & 44.10 & 45.02 & 57.12 & 56.20 & 44.10 & 45.02 \\\hline
\end{tabular}
}
\end{table}
\endgroup
\begingroup
\setlength{\tabcolsep}{10pt}
\begin{table}[t]
\centering
\caption{\textbf{Code-specific metrics.} We present the AST accuracy and CodeBLEU score of models under multi-step planning with code generation with or without feedback.}
\label{tab:code-metrics}
\resizebox{\columnwidth}{!}{%
\begin{tabular}{lllllllll}
                   & \multicolumn{4}{l}{AST accuracy} & \multicolumn{4}{l}{CodeBLEU}  \\\hline \hline
model              & P      & PV     & PE     & PVE   & P     & PV    & PE    & PVE   \\ \hline
Llama-2-7b         & 0.00   & 0.00   & 0.00   & 0.00  & 22.64 & 21.28 & 17.58 & 21.19 \\
Llama-2-13b        & 0.11   & 0.23   & 0.00   & 0.00  & 29.96 & 27.09 & 20.29 & 27.62 \\
Mixtral-8x7B       & 2.04   &  3.06      &   4.22     & 2.30      & 54.17 &   48.48    &   53.01    &  47.21     \\
Gemini-pro         & 3.85   & 5.33   & 3.74   & 4.54  & 62.37 & 61.13 & 59.00 & 59.18 \\
GPT-3.5-turbo-0125 & 3.29   & 4.76   & 3.29   & 4.42  & 60.79 & 60.32 & 58.96 & 59.99 \\
GPT-4-0125-preview & 4.31   & 5.10   & 4.42   & 5.33  & 68.52 & 68.37 & 68.68 & 68.51\\ \hline
\end{tabular}
}
\end{table}
\endgroup
\begingroup
\setlength{\tabcolsep}{10pt}
\begin{table}[t]
\centering
\caption{\textbf{Average turn count.} We present the average number of conversation turns in step-by-step and multi-step planning with JSON-format generation and different types of feedback.}
\label{tab:turn-count}
\resizebox{0.8\columnwidth}{!}{%
\begin{tabular}{lllllll}
                                    &              & \multicolumn{5}{l}{Average \# of turns} \\\hline \hline
model                               & strategy     & N/A     & P      & PV     & PE     & PVE    \\\hline
\multirow{2}{*}{Llama-2-7b}         & step-by-step & 2.00    & 3.54   & 4.03   & 3.26   & 3.52   \\
                                    & multi-step   & 1.00    & 1.10   & 2.18   & 1.95   & 1.99   \\
\multirow{2}{*}{Llama-2-13b}        & step-by-step & 2.87    & 2.87   & 3.09   & 3.06   & 2.99   \\
                                    & multi-step   & 1.00    & 1.04   & 1.98   & 1.91   & 1.97   \\
\multirow{2}{*}{Mixtral-8x7B}       & step-by-step & 2.98    & 6.37   & 5.55   & 6.02   & 6.09   \\
                                    & multi-step   & 1.00    & 1.14   & 2.43   & 2.74   & 2.81   \\
\multirow{2}{*}{Gemini-pro}         & step-by-step & 2.31    & 3.01   & 2.28   & 3.67   & 3.78   \\
                                    & multi-step   & 1.00    & 1.20   & 1.84   & 1.80   & 1.88   \\
\multirow{2}{*}{GPT-3.5-turbo-0125} & step-by-step & 2.40    & 3.39   & 4.10   & 5.43   & 5.30   \\
                                    & multi-step   & 1.00    & 1.02   & 1.36   & 1.46   & 1.62   \\
\multirow{2}{*}{GPT-4-0125-preview} & step-by-step & 3.22    & 3.52   & 3.51   & 3.59   & 3.59   \\
                                    & multi-step   & 1.00    & 1.00   & 1.05   & 1.06   & 1.07  \\\hline
\end{tabular}
}
\end{table}
\endgroup
\begingroup
\setlength{\tabcolsep}{4pt}
\begin{table}[t]
\caption{\textbf{Average number of input and output tokens}}
\label{tab:token-count}
\resizebox{\columnwidth}{!}{%
\begin{tabular}{llrrrrr|rrrrr}
                                    &              & \multicolumn{5}{l}{Avg \# of input tokens}           & \multicolumn{5}{|l}{Avg \# of output tokens} \\\hline\hline
model                               & strategy     & N/A      & P        & PV       & PE       & PVE      & N/A     & P      & PV     & PE     & PVE    \\ \hline
\multirow{2}{*}{Llama-2-7b}         & step-by-step & 5497.25  & 20627.60 & 22021.08 & 14356.79 & 13562.25 & 108.54  & 659.02 & 673.01 & 436.63 & 432.34 \\
                                    & multi-step   & 2184.19  & 3065.88  & 10215.74 & 6792.83  & 8570.81  & 273.65  & 320.95 & 735.02 & 478.79 & 636.73 \\
\multirow{2}{*}{Llama-2-13b}        & step-by-step & 13084.77 & 14793.73 & 13962.84 & 11498.10 & 13025.18 & 535.74  & 620.00 & 495.34 & 446.56 & 489.17 \\
                                    & multi-step   & 2184.19  & 2651.22  & 8141.48  & 7375.54  & 8309.38  & 326.91  & 345.01 & 738.19 & 648.41 & 753.93 \\
\multirow{2}{*}{Gemini-pro}         & step-by-step & 5661.28  & 7651.78  & 5653.98  & 10136.36 & 10560.46 & 115.70  & 171.22 & 96.98  & 216.03 & 232.53 \\
                                    & multi-step   & 2184.19  & 3062.00  & 4962.19  & 4786.80  & 5022.53  & 86.12   & 155.05 & 219.64 & 216.77 & 225.45 \\
\multirow{2}{*}{GPT-3.5-turbo-0125} & step-by-step & 5891.36  & 8938.04  & 11693.37 & 16497.09 & 15966.33 & 109.61  & 189.53 & 207.51 & 317.43 & 318.30 \\
                                    & multi-step   & 2184.19  & 2247.54  & 3199.10  & 3502.05  & 4017.90  & 96.24   & 99.47  & 136.24 & 149.94 & 166.76 \\
\multirow{2}{*}{GPT-4-0125-preview} & step-by-step & 8046.55  & 8852.87  & 8832.17  & 9601.61  & 9618.19  & 166.17  & 172.37 & 171.03 & 235.51 & 236.76 \\
                                    & multi-step   & 2184.19  & 2184.19  & 2318.98  & 2331.06  & 2354.78  & 102.28  & 103.49 & 110.55 & 107.74 & 111.09 \\\hline
\end{tabular}
}
\end{table}
\endgroup

Apart from the three main metrics in the main paper, we have also evaluated all six large language models on 10+ other metrics. We report these additional evaluation results below. 
\subsection{No feedback} In the main paper, we present the results of models with verification and/or execution of feedback (on top of parsing feedback) using the experiment with parsing (P) feedback as a baseline. Here, we report the results using the experiment with no feedback at all as the baseline in Table \ref{tab:no-feedback}. We see that our main takeaway remains the same with this change: feedback helps improve models' argname-F1 by a small amount and pass rate by a lot, although it can lead to a small decrease in tool-F1. We additionally observe the improvement of verification and/or execution feedback on pass rate is larger than that of parsing feedback. 

\subsection{Step-level metrics} Besides tool-F1 and argname-F1, we also report the following step-level metrics: argvalue-F1 (Table \ref{tab:argvalue-f1}), edge-F1 (Table \ref{tab:edge-f1}), and normalized edit distance (Table \ref{tab:norm_edit_dist}). We adapted TaskBench's \cite{shen2023taskbench} implementation of these metrics on our benchmark.  We caution readers about argvalue-F1 as it is computed based on exact matching to one groundtruth value even though there can be multiple valid values. 

\subsection{Plan-level accuracy} Since step-level metrics do not take into account the ordering of the predicted tools, we additionally include plan-level accuracy to evaluate the whole plan's correctness (Table \ref{tab:plan-acc}). We highlight two main variants of plan accuracy in Table \ref{tab:plan-acc}, where the first one considers a list of tool names as a plan and the second considers a list of (tool name, argument names) tuples as a plan. As there could be multiple valid plans of the same query, we have also included the $\Delta$ in plan accuracy considering alternative plans in Table \ref{tab:plan-acc-alt} and shown that our set of alternative plans can recover 1-5\% examples where the models could have output potential valid plans different from the one human-verified groundtruth plan. Finally, we also present the strictest form of plan accuracy, which considers a list of tool names, argument names and values as a plan in Table \ref{tab:plan-acc-argvalue}. We note that exact matching gives us (Table \ref{tab:plan-acc-argvalue} left) extremely low scores while using entailment in the case of text values -- if the predicted argument text entails the label text -- gives us more reasonable scores (Table \ref{tab:plan-acc-argvalue} right). 

Additionally, we also include the plan accuracy of models across different numbers of tools with multi-step and step-by-step planning respectively in
Tables \ref{tab:plan-acc-breakdown} and \ref{tab:plan-acc-breakdown-local}. Under multi-step planning, we find that most models experience a drop in plan accuracy as the number of tools in the plans increases. Interestingly, the smaller models like LLama-7b and 13b exhibit a slightly different trend, achieving a higher number on 2-tool examples than on 1-tool ones (Table \ref{tab:plan-acc-breakdown}. One plausible explanation is that these models might not fully understand the user request and tend to output 2-tool plans more often. Surprisingly, GPT-4 also scores higher on 2-tools examples than 1-tool ones but with a much smaller gap.  On the other hand, under step-by-step planning, we see that all models suffer from an even more drastic drop in plan accuracy as the number of tools required increases (Table \ref{tab:plan-acc-breakdown-local}). \textbf{This suggests that step-by-step planning might not scale well to more complex tasks that require a large number of tools/actions.}

\subsection{Code-specific metrics: AST accuracy and CodeBLEU}
To evaluate code generation properly, we have also included code-specific metrics such as AST accuracy and CodeBLEU (Table \ref{tab:code-metrics}). AST accuracy measures if the AST tree of the predicted code is the same as the label code, whereas CodeBLEU measures the similarity of the predicted code to the reference code. We find that feedback, especially verification feedback, can help improve models' AST accuracy but not necessarily CodeBLEU scores. 

\subsection{Efficiency}
Besides models' planning performance, we also kept track of their token usage (Table \ref{tab:token-count}) and numbers of conversation turns (Table \ref{tab:turn-count}). As expected, step-by-step planning generally requires more conversation turns and more tokens than multi-step planning. Similarly, feedback also increases token usage.

\section{Evaluation of plan execution outputs}
\subsection{Human evaluation}
\begingroup
\setlength{\tabcolsep}{10pt}
\begin{table}[t]
\centering
\caption{\textbf{Human evaluation of execution outputs.} We present the execution accuracy of GPT-4 and Mixtral-8x7B on a selected subset of 85 examples across different setups, including step-by-step and multi-step planning, with JSON-format and code generation, and different types of feedback.}
\label{tab:exec-acc}
\resizebox{0.8\columnwidth}{!}{%
\begin{tabular}{llllr}
model              & strategy     & format & feedback & accuracy    \\\hline\hline
Mixtral-8x7B       & multi-step   & JSON   & P        & $42.94 \pm 1.76$ \\
GPT-4-0125-preview & step-by-step & JSON   & P        & $49.41 \pm 1.18$ \\
GPT-4-0125-preview & multi-step   & Code   & P        & $61.18 \pm 0.0$  \\
GPT-4-0125-preview & multi-step   & JSON   & PVE      & $64.12 \pm 2.94$ \\
GPT-4-0125-preview & multi-step   & JSON   & P        & $70.00 \pm 6.47$\\\hline
\end{tabular}
}
\end{table}
\endgroup
\begingroup
\setlength{\tabcolsep}{10pt}
\begin{table}[t]
\centering
\caption{\textbf{Automatic evaluation of execution outputs.} We present the execution  accuracy of models in step-by-step and multi-step planning and with JSON-format and code generation.}
\label{tab:exec-acc-auto}
\resizebox{0.8\columnwidth}{!}{%
\begin{tabular}{llll}
model                               & strategy     & format & accuracy \\ \hline \hline
\multirow{3}{*}{Llama-2-7b}         & step-by-step & JSON   & 10.52    \\
                                    & multi-step   & JSON   & 7.54     \\
                                    & multi-step   & code   & 1.45     \\ \hline
\multirow{3}{*}{Llama-2-13b}        & step-by-step & JSON   & 9.17     \\
                                    & multi-step   & JSON   & 12.27    \\
                                    & multi-step   & code   & 7.43     \\ \hline
\multirow{3}{*}{Mixtral-8x7B}       & step-by-step & JSON   & 34.16    \\
                                    & multi-step   & JSON   & 42.28    \\
                                    & multi-step   & code   & 34.03    \\ \hline
\multirow{3}{*}{Gemini-pro}         & step-by-step & JSON   & 45.74    \\
                                    & multi-step   & JSON   & 44.25    \\
                                    & multi-step   & code   & 34.28    \\ \hline
\multirow{3}{*}{GPT-3.5-turbo-0125} & step-by-step & JSON   & 38.36    \\
                                    & multi-step   & JSON   & 49.47    \\
                                    & multi-step   & code   & 42.85    \\ \hline
\multirow{3}{*}{GPT-4-0125-preview} & step-by-step & JSON   & 50.86    \\
                                    & multi-step   & JSON   & 60.51    \\
                                    & multi-step   & code   & 54.49  \\ \hline 
\end{tabular}}
\end{table}
\begingroup
\setlength{\tabcolsep}{10pt}
\begin{table}[t]
\centering
\caption{\textbf{Automatic evaluation of execution outputs with feedback.} We present the execution accuracy of models in multi-step planning with various feedback.}
\label{tab:exec-acc-feedback}
\resizebox{0.8\columnwidth}{!}{%
\begin{tabular}{lrrrr}
model              & P     & PV    & PE    & PVE   \\ \hline \hline
Llama-2-7b         & 7.54  & 13.54 & 8.90  & 8.51  \\
Llama-2-13b        & 12.27 & 25.44 & 22.58 & 16.29 \\
Mixtral-8x7B       & 42.28 & 45.57 & 46.19 & 42.92 \\
Gemini-pro         & 44.25 & 51.72 & 49.47 & 55.18 \\
GPT-3.5-turbo-0125 & 49.47 & 55.23 & 55.12 & 55.26 \\
GPT-4-0125-preview & 60.51 & 60.72 & 59.07 & 61.73 \\ \hline
\end{tabular}}
\end{table}
Since \name consists of open-ended queries, which do not always have one single final answer, it is challenging to evaluate the execution results of the plans automatically. Thus, we resort to human evaluation of a small subset of 85 examples with reasonable execution results. Our manual evaluation reveals that GPT-4 achieves the best execution accuracy with multi-step planning and JSON-format generation compared to step-by-step planning or code generation (Table \ref{tab:exec-acc}). 

\subsection{Automatic evaluation}
Nevertheless, as human evaluation is not scalable, we have also implemented automatic evaluation, which uses the groundtruth plans' final outputs as the golden answers and compares the predicted plans' results against them. Our implementation invokes different metrics (all in [0,1]) based on the outputs' modality: cosine similarity with SentenceBERT embeddings between texts, and CLIP embeddings between images and Average Precision for predicted objects.  We report the average accuracy on 210 queries with plans that yield good and deterministic outputs in Tables \ref{tab:exec-acc-auto} and \ref{tab:exec-acc-feedback}. 

Similar to our planning evaluation, most models achieve the best execution accuracy in multi-step planning with JSON format generation except for LLama-7b and Gemini-pro, where step-by-step planning leads to higher execution accuracy (Table \ref{tab:exec-acc-auto}). In addition, we also observe that verification and/or execution feedback do lead to some improvement (up to 10+\%) in execution accuracy compared to parsing feedback only, with only one exception in GPT-4 with execution feedback where the execution accuracy is comparable but not better (Table \ref{tab:exec-acc-feedback}). Overall, these results suggest our execution output evaluation aligns well with our planning evaluation, providing further evidence to support our findings on the positive effects of multi-step planning, JSON-format generation and feedback.

\clearpage


%
%

\bibliographystyle{splncs04}
\bibliography{main}
\end{document}